\documentclass[final]{cvpr}

\usepackage{times}
\usepackage{epsfig}
\usepackage{graphicx}
\usepackage{amsmath}
\usepackage{amssymb}
\usepackage{xcolor}
\usepackage{tabularx}
\usepackage{bbm}
\usepackage{capt-of}

\usepackage[pagebackref=true,breaklinks=true,colorlinks,bookmarks=false]{hyperref}

\newcommand{\tali}[1]{}
\newcommand{\miki}[1]{}
\newcommand{\fcole}[1]{}
\newcommand{\erika}[1]{}
\newcommand{\az}[1]{}
\newcommand{\todo}[1]{}

\newcommand{\el}{\etal}

\def\cvprPaperID{344} %
\def\confYear{CVPR 2021}

\newcommand{\afterfigure}{\vspace{-1em}}

\begin{document}

\title{\vspace{-.2in} Omnimatte: Associating Objects and Their Effects in Video}
\author{Erika Lu$^{1,2}$ \qquad
Forrester Cole$^{1}$
\qquad
Tali Dekel$^{1,3}$\qquad
Andrew Zisserman$^{2}$\\ \vspace{0.4cm}
William T. Freeman$^{1}$
\qquad
Michael Rubinstein$^{1}$\\
{$^1$Google Research \qquad \qquad}  $^2$University of Oxford \qquad \qquad $^3$Weizmann Institute of Science
\\
}

\twocolumn[{%
\renewcommand\twocolumn[1][]{#1}%
\maketitle
\vspace*{-0.3cm}
\includegraphics[width=1\textwidth]{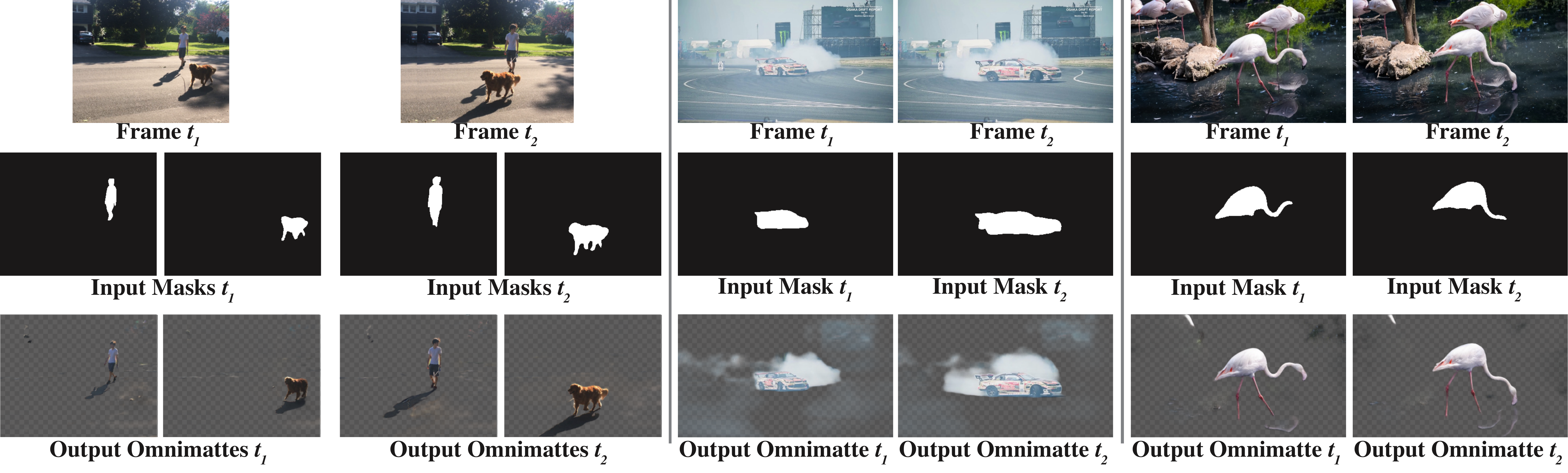}
\captionof{figure}{ We pose a novel problem: automatically associating subjects
in videos with `effects' related to them in the scene. Given an input video (top) and rough masks of subjects of interest (middle), our method estimates an \emph{omnimatte} -- an alpha matte and foreground color that includes the subject itself along with all scene elements associated with it (bottom). The associated elements can be other objects attached to the subject or moving with it, or complex effects such as shadows, reflections, smoke, or ripples the subject creates in water. \vspace*{0.6cm}}
\label{fig:teaser}
}]

\maketitle
\pagenumbering{gobble}
\begin{abstract}
\vspace{-.1in}
Computer vision is increasingly effective at segmenting objects in images and videos; however, scene effects related to the objects---shadows, reflections, generated smoke, \etc---are typically overlooked. Identifying such scene effects and associating them with the objects producing them is important for improving our fundamental understanding of visual scenes, and can also assist a variety of applications such as removing, duplicating, or enhancing objects in video. In this work, we take a step towards solving this novel problem of automatically associating objects with their effects in video. Given an ordinary video and a rough segmentation mask over time of one or more subjects of interest, we estimate an \emph{omnimatte} for each subject---an alpha matte and color image that includes the subject along with all its related time-varying scene elements. Our model is trained only on the input video in a self-supervised manner, without any manual labels, and is generic---it produces omnimattes automatically for arbitrary objects and a variety of effects. We show results on real-world videos containing interactions between different types of subjects (cars, animals, people) and complex effects, ranging from semi-transparent elements such as smoke and reflections, to fully opaque effects such as objects attached to the subject.\footnote{Project page: \url{https://omnimatte.github.io/}
}

\end{abstract}

\vspace{-.2cm}
\section{Introduction}
\vspace{-.1cm}
\textit{``And first he will see the shadows best, next the reflections of men and other objects in the water, and then the objects themselves, then he will gaze upon the light of the moon and the stars and the spangled heaven ... Last of all he will be able to see the sun."} -- Plato

\begin{figure*}
\vspace{-.3in}
    \centering
    \includegraphics[width=0.85\textwidth]{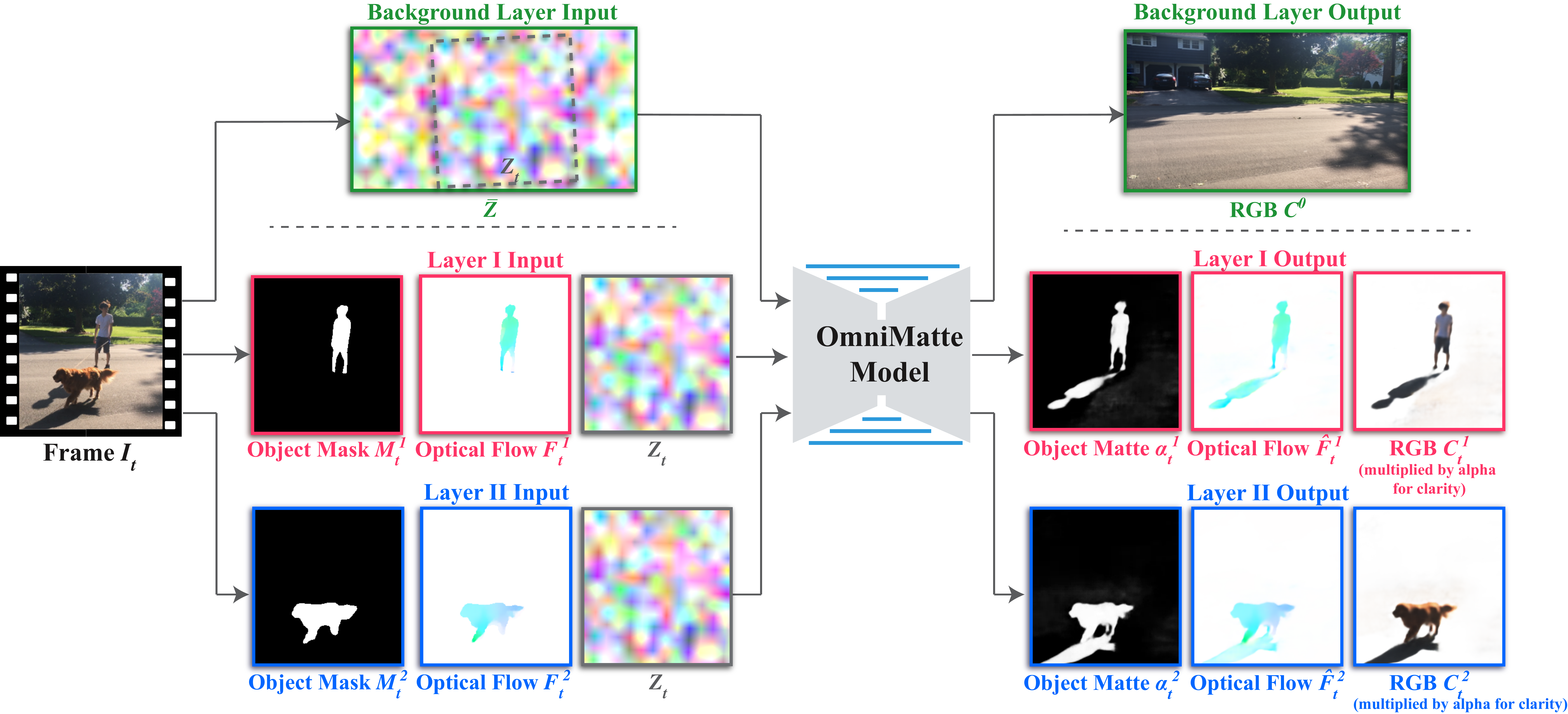}
    \caption{{\bf Estimating omnimattes from video.} The input to the model is an ordinary video with multiple moving objects, and a rough segmentation mask $M$ for each object (left). In a pre-processing step, we compute an optical flow field $F$ between consecutive frames using~\protect\cite{teed2020raft}. For each object, we pass the mask, estimated flow in the object's region, and a sampled noise image $Z_t$ (representing the background) to our model, producing an omnimatte (color + opacity) and an optical flow field for the object (right). In addition, the model predicts a single background color image for the entire video (top), given a spatial texture noise image $\bar{Z}$ as input. See Sec.~\ref{sec:method} for details.}
    \label{fig:pipeline}
\afterfigure    
\end{figure*}

Is it possible to automatically determine all the effects caused by a subject in a video? Reflect for a moment on the difficulty of the task:
a subject, such as a human wandering through a scene, can cast shadows on the floor and distant walls, and be reflected in windows and other surfaces. These `effects' are non-local. However, they are {\em correlated} with the subject's shape, motion and, in the case of reflections, appearance.

Tackling this problem is the objective of this paper. More specifically, given an input video and (possibly rough) segmentations over time of subjects of interest in the video, we seek to produce an output opacity matte (alpha matte) for each subject
that includes the subject and their effects in the scene (Figure~\ref{fig:teaser}). We call this the \emph{``omnimatte''} of the subject. We additionally produce a color background image containing the static background elements in the video. We achieve this by proposing a network and training framework that is able to automatically determine and segment regions that are {\em correlated} with the given subject (Figure~\ref{fig:pipeline}). The model is trained in a self-supervised way only on the input video, without observing any additional examples. Our solution is inspired by the recent work of Lu \etal~\cite{lu2020layered} that presented a method to decompose a video into a set of human-specific RGBA layers. We generalize this technique to support arbitrary objects, by relying only on binary input masks (no object-specific representation or processing) and incorporating general optical flow to account for motion and frame-to-frame correspondence.

Associating objects with their effects not only improves our fundamental understanding of visual scenes and events captured in video, it can also support a range of applications. Consider for example the problem of removing a person or other types of objects from a video. As is well known, a common error in person removal, e.g., by inpainting, is that a shadow or reflection of the person remains, resulting in a video left with just a `shadow of the former self'. The erroneous missing of a reflection is a central plot point in the film `Rising Sun' (1993), and the converse, a lack of reflection, a common trope of vampire movies. The important point is that manipulating an object in a video requires dealing not only with the object; its \emph{effects} in the scene need to be adjusted together with the object in order to create realistic and faithful renditions.

We demonstrate results of inferring omnimattes for different objects such as animals, cars, and people, capturing a variety of complex scene effects including shadows, reflections, dust and smoke. We evaluate the resulting omnimattes qualitatively and quantitatively, and also demonstrate how omnimattes can be useful for video editing applications such as object removal, background replacement, ``color pop'', and stroboscopic photography.

\section{Related Work}

\paragraph{Video layer decomposition}  Our work is inspired by seminal works on layered video decomposition such as~\cite{wangadelson1994} and~\cite{brostow1999motion}. Layered representations of images and videos have been applied widely in computer vision and graphics, for example, for inferring occlusion relationships (\eg~\cite{brostow1999motion}), depth (\eg~\cite{zhou2018stereo,srinivasan2019}), mosaics (\eg~\cite{Fradet08}), and synthesizing novel views (\eg~\cite{shade1998layered,tulsiani2018layer}). In particular, our work builds on the recent work of Lu~\el~\cite{lu2020layered} that presented a method for decomposing a video into a set of human-specific RGBA layers, where each layer represents a person and their associated scene elements. They take a neural rendering approach and represent the geometry and texture of people explicitly, using a dedicated, human-specific pipeline. We demonstrate that the neural rendering component is in fact unnecessary, and that comparable results can be achieved by providing only rough segmentation masks and flow as input to the network. The result is a simpler and more efficient setup that allows the model to handle arbitrary moving objects. We compare the results with~\cite{lu2020layered} in Sec.~\ref{sec:retiming}.

\vspace{-1em}
\paragraph{Image and video matting}
Image and video matting traditionally deals with the problem of estimating a foreground layer (color + opacity) and a background color image from a given image or a video (e.g.,~\cite{bai09,wang05,li05,chuang02vidmat,xu17,hou2019context,BMSengupta20}). The novel problem we propose---estimating an omnimatte---also aims at estimating color + opacity layers from an input video. However, the key fundamental difference is that omnimattes capture not only an object but also all the various scene effects that are correlated with the object. None of the existing matting methods is suitable for performing this task: they cannot handle well entirely semi transparent objects, typically require accurate trimaps that are generated manually, and they are often restricted to estimating two layers (background/foreground). In practice, film production uses manual or semi-automatic rotoscoping to create mattes with such effects~\cite{li2016roto++}. Our method works automatically and generically on natural, ordinary videos that contain arbitrary moving objects and scene effects, and requires only rough object masks (e.g., see the flamingo example in Fig.~\ref{fig:teaser}). 

\vspace{-1em}
\paragraph{Background subtraction}
Change detection using background subtraction~\cite{piccardi2004background,elgammal2000non} typically does not produce alpha mattes, but  binary masks containing all objects and effects (such as shadows). Qian and Sezan's ``difference matting'' work~\cite{qian1999video} attempts to use background subtraction and thresholding to produce a foreground matte with a known background image, but the results are very sensitive to the thresholding value. \cite{BMSengupta20} recently modernized that approach, producing nice quality, continuous-valued alpha mattes but still require a known, clean image of the background. More importantly, background subtraction and difference matting cannot solve the omnimatte problem when the video has \emph{multiple objects with effects}. In such cases it is not enough to detect the effects, each effect must also be associated its subject. We evaluate our method numerically and compare it with background subtraction using a change detection dataset~\cite{wang14cdw} with pixel-level labels for objects and shadows.

\vspace{-1em}
\paragraph{Shadow and reflection detection}
Specialized methods also exist for detecting, removing, or modifying specific types of effects, such as shadows and reflections. 
For example, \cite{chuang2003shadow} acquire a shadow displacement map by waving a stick over different parts of a scene, then use it to synthesize realistic shadows that match an object's shape. \cite{alayrac2019, alayrac2019ICCV} decompose natural videos to remove reflections, shadows, and smoke. More recently, Wang~\el~\cite{WangCS20} proposed to analyze the motion of people in video and use it to predict depth, occlusion, and lighting/shadow information, to increase realism of 2D object insertion.
Our goal in this work is to provide a general technique for inferring all of a subject's associated effects. However, as shadows are a particularly common effect, we compare our results with a state-of-the-art shadow detector~\cite{wang2020isd} in Sec.~\ref{sec:results_qual}.

\vspace{-1em}
\paragraph{Video Effects}
Although omnimattes are not explicitly optimized for editing, they can facilitate various video editing effects that rely on input object masks, including object removal and video completion (\eg~\cite{Gao-ECCV-FGVC,wexler07,Newson14}), object cut-and-paste (\eg~\cite{li05,wang05}), color pop, and creation of stroboscopic images from video (\eg~\cite{agarwala2004interactive}). All of these effects can be achieved via simple manipulations of the estimated omnimattes in a post-processing step, or alternatively, by using the omnimattes as input to existing methods such as video completion (\eg~\cite{Gao-ECCV-FGVC}), to save the manual work required for marking the object's effects. We demonstrate these results in Sec.~\ref{sec:results_qual}.

\section{Estimating Omnimattes from Video}
\label{sec:method}
The input to our method is an ordinary video of moving objects, and one or more layers of rough segmentation masks that mark the subjects of interest.
The output is an \emph{omnimatte} for each input mask layer, consisting of an alpha matte (opacity map) and a color image. 
The model is trained per-video to reconstruct the input in a self-supervised manner, without observing any additional examples.

To accurately reconstruct the input video, the model must infer all the time-varying effects (\eg shadows, reflections) from the input object masks, which do \textit{not} represent those effects.
Our goal is to steer the model to place the associated effects in the layer of the subject causing them. 
Lu, \etal~\cite{lu2020layered} showed that this association can be achieved by 
showing the network one mask at a time, leveraging the fact that an effect is easier to predict from the object mask most correlated with it.
For example, the mask of the person in Figure~\ref{fig:teaser} provides more information about its shadow (more similar to it in shape, in motion) compared to the mask of the dog. Therefore (as shown in~\cite{lu2020layered}) the network tends to learn to predict the person's shadow from the person's mask (thus associating it with the correct layer). We build on this training strategy, but design network inputs and losses to encourage this solution for general objects.

\subsection{Overview}
\label{sec:training_overview}
Figure~\ref{fig:pipeline} illustrates our pipeline.  Our model is a 2D U-Net~\cite{ronneberger2015unet} that processes the video frame by frame.  For each frame, we compute rough object masks using off-the-shelf techniques to mark the major moving objects in the scene. We group the objects into $N$ mask layers $\{M^i_t\}_{i=1}^N$ and define a (possibly time-varying) ordering $o_t$ for the layers. For example, in a scene with a rider, a bicycle, and several people in a crowd, we might group the rider and bicycle into one layer, while grouping the crowd into a second layer. To equip our model with explicit information about \emph{object motion} and frame-to-frame correspondence, we also compute a dense optical flow field, $F_t$, between each frame and the consecutive frame in the video. This flow field is masked by the input masks $M^i_t$ to provide the network only flow information related to the layer's subject.
We additionally align all frames onto a common coordinate system using homographies, and represent the background as a single unwrapped image on a separate layer.

From this rough yet explicit representation of moving objects, the model has to infer: (i) \emph{omnimattes} -- pairs of continuous-valued opacity maps (mattes) and RGB images that capture not only the $i^{th}$ moving object but also all the scene elements that are correlated with it in space and time (e.g., reflections, shadows, attached objects, etc.), (ii) a refined optical flow field for each layer, and (iii) a background RGB image. Formally,
{\small 
\begin{equation}
 \text{Omnimatte}(I_t, H_t, M_t^i, F_t^{i}) = \mathcal{L}_t = \{\alpha_t^i, C_t^i, \hat{F}^{i}_t\},
\end{equation}} 
where $I_t$, $H_t$, $M_t^i$, $F_t^i$ are the input video RGB frame, estimated camera homography, the initial input mask, and the pre-computed flow field of the $i^{th}$ object in time $t$, respectively. $\alpha^i_t$ and $C_t^i$ are the alpha and color buffers of the output omnimatte, and $\hat{F}^i_{t}$ is the predicted object flow. 

The training loss consists of terms on the RGBA outputs (Sec.~\ref{sec:rgba_losses}) and the predicted flow (Sec.~\ref{sec:flow_losses}). The main loss is a reconstruction loss $\mathbf{E}_\text{rgb-recon}$, but as reconstruction is underconstrained with multiple layers, we add a sparsity regularization $\mathbf{E}_\text{reg}$ to the alpha layers and an initialization loss $\mathbf{E}_\text{mask}$ to the masks, similar to~\cite{lu2020layered}. We encourage the \emph{motion} of the result to match the input by adding a flow-reconstruction loss $\mathbf{E}_\text{flow-recon}$ and a temporal consistency term to the alpha mattes $\mathbf{E}_\text{alpha-warp}$. 

The total loss is:
{\small 
\begin{equation}
    \mathbf{E}_\text{rgb-recon} + \lambda_\text{r} \mathbf{E}_\text{reg} +
    \lambda_\text{m} \mathbf{E}_\text{mask} + \\
    \mathbf{E}_\text{flow-recon} +  \\
    \lambda_\text{w} \mathbf{E}_\text{alpha-warp},
    \label{eq:total_obj}
\end{equation}}
where $\lambda_\text{r}$, $\lambda_\text{m}$, and $\lambda_\text{w}$ are weighting coefficients (see supplementary material (SM)).
As the background is assumed to be static, we factor out camera motion and treat the background with a special, fixed layer (Sec.~\ref{sec:camera}).

\subsection{RGBA Losses}
\label{sec:rgba_losses}
The main loss in our optimization is a \emph{reconstruction loss}. Formally, we composite the set of estimated layers for each frame and  the predicted background layer using standard back-to-front compositing~\cite{porterduff1984}, and encourage the composite image to match the original frame:
{\small \begin{equation}
    \mathbf{E}_\text{rgb-recon} = \frac{1}{T}\sum_t \|I_t  - Comp(\mathcal{L}_t, o_t)\|_1,
    \label{eq:rgb_recon}
\end{equation}}
where $\mathcal{L}_t  = \{\alpha^i_t, C_t^i\}_{i=1}^N$ are the predicted layers for frame $t$, and $o_t$ is the compositing order.

To prevent a trivial solution where a single layer reconstructs the entire frame, we further apply a regularization loss to the $\alpha_t^i$ to encourage them to be spatially sparse. We use a mix of $L_1$ and an approximate-$L_0$:
{\small \begin{equation}
    \mathbf{E_{\text{reg}}}  =  \frac{1}{T}\frac{1}{N}\sum_t \sum_i  \gamma \left \| \alpha_t^i \right \|_1 + \Phi_0(\alpha_t^i),
    \label{eq:matte_reg}
\end{equation}}
where $\Phi_0(x) =  2\cdot \mathtt{Sigmoid}(5x) - 1$ smoothly penalizes non-zero values of the alpha map, and $\gamma$ controls the relative weight between the terms. 

To guide the optimization to convergence from a random initialization, we therefore adopt a ``bootstrap'' loss to coerce the alpha maps $\alpha_t^i$ to match the input masks $M_t^i$:
{\small \begin{equation}
    \mathbf{E}_\text{mask} = \frac{1}{T}\frac{1}{N}\sum_t \sum_i \left \| d_t^i \odot (M_t^i - \alpha_t^i) \right \|_2
\end{equation}}
where $d_t^i = 1 - \mathtt{dilate}(M_t^i) + M_t^i$ is a boundary erosion mask to turn off the loss near the mask boundary, and $\odot$ is element-wise product. This loss is turned off after its value reaches a fixed threshold (see SM). 

\subsection{Flow Losses}
\label{sec:flow_losses}

Our model additionally predicts a set of \emph{flow layers}. Predicting flow layers serves as an auxiliary task that injects information about motion to our model and improves our decomposition (as demonstrated by our experiments). To achieve that we apply a \emph{flow reconstruction loss} and a \emph{photometric warping loss} defined below:
{\small \begin{equation}
    \mathbf{E}_\text{flow-recon} = \frac{1}{T}\sum_t W_t \cdot \|F_t  - \textit{Comp}(\mathcal{F}_t, o_t)\|_1,
    \label{eq:flow_recon}
\end{equation}}
where $\mathcal{F}_t = \{\hat{F}_i^t\}$ is the set of predicted flow layers, $F_t$ is the original, pre-computed flow, and $W_t$ is a spatial weighting map that lowers the impact of pixels with inaccurate flow. $W_t$ is computed based on standard left-right flow consistency error and photometric warping error (see full details in SM).

We additionally encourage temporal consistency within layers using an \textit{alpha warping loss}:
{\small \begin{equation}
    \mathbf{E}_\text{alpha-warp} = \frac{1}{T}\frac{1}{N}\sum_t \sum_i \|\alpha_t^i  - \alpha_{wt}^i\|_1,
    \label{eq:warpping}
\end{equation}}
where $\alpha_{wt}^i = \textit{Warp}(\alpha_{t+1}^i, \mathcal{F}_t^i)$ is the alpha for layer $i$ at time $t+1$ warped to time $t$ using the predicted flow.

\subsection{Camera Motion and Background}
\label{sec:camera}
We assume the background scene is stationary and camera motion can be modeled by a time-varying homography from an unwrapped ``canvas'' image, as in \cite{wangadelson1994}. 
The homographies $H_t$ from frame $t$ to the canvas are estimated via feature tracking (using~\cite{grundmann2011auto}) on the original RGB video frames and are held fixed. For input to the network, the background canvas is represented by a single spatial noise image $\Bar{Z}$ (see Fig.~\ref{fig:pipeline}). The background color layers $C_t^0$ are produced by feeding $\Bar{Z}$ through the network to form a static color image $\Bar{C^0}$, which is then sampled using $H_t^{-1}$ to form time-varying background images $\{C_t^0\}$.

To make the foreground layers aware of the camera motion, the input mask layers $M_t^i$ are concatenated with a noise image that tracks the camera. The background noise image $\Bar{Z}$ is sampled using $H_t^{-1}$ to form time-varying noise images $\{Z_t\}$. 
This is a similar approach to Lu, et al.~\cite{lu2020layered}, but in our case the noise image is not trainable. 

Minor stabilization errors, as well as exposure changes, vignetting, and radial distortion, usually cause slight changes in appearance even for a stationary background. If the background is assumed to be entirely static, these subtle shifts in appearance will show up as noise in our omnimatte. Such effects, however, tend to have low spatial and temporal frequency relative to the subject's effects and can be safely captured by applying a \emph{refinement warp} to the background layer. The refinement warp consists of a spatially and temporally coarse grid-based warp. We additionally apply a grid-based brightness adjustment to the final composite $Comp(\mathcal{L}_t, o_t)$. The parameters of the warp and brightness adjustment are optimized together with the network parameters (see SM for additional details).

\subsection{Implementation Details}
For all our results, we used Mask R-CNN~\cite{he17} to segment the input objects, and STM~\cite{stm}
(a video object segmenter trained on the DAVIS dataset~\cite{davis2017}) to track objects across frames.
Optical flow between consecutive frames was computed using RAFT~\cite{teed2020raft}.
When dynamic background elements such as tree branches are present, we use panoptic segmentation~\cite{wu2019detectron2} to segment them and treat the segment as additional objects. 
To increase the detail of the color buffers $C_t^i$, we apply a similar detail-transfer technique to Lu, \etal~\cite{lu2020layered}. 
See SM for training details.

\section{Results} 
\begin{figure*}[t]
\vspace{-.2in}
    \centering
    \includegraphics[width=\textwidth]{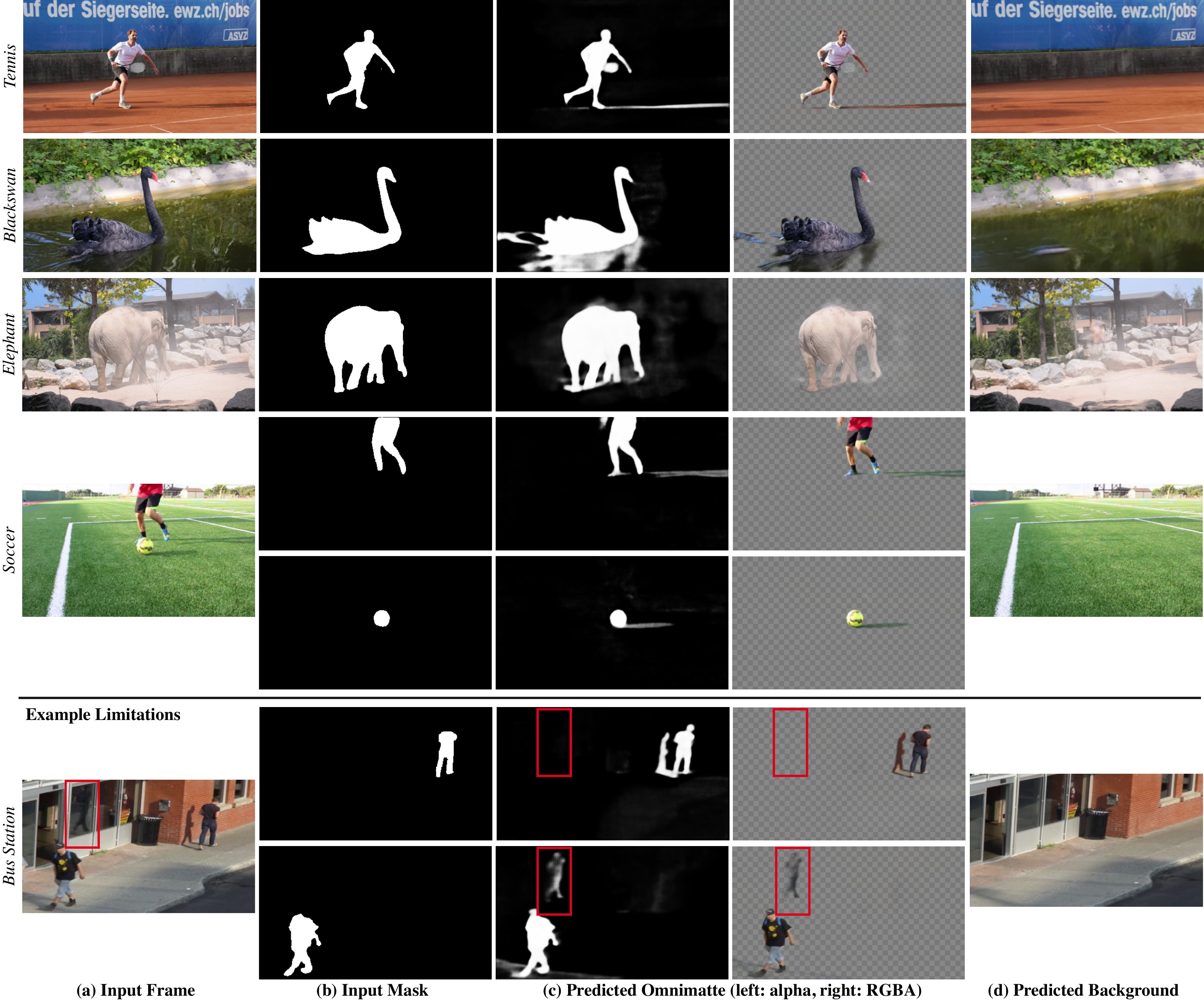}
    \vspace{-.2in}
    \caption{{\bf Results on natural videos.} For each example, we show: \textbf{(a)} input frame; \textbf{(b)} input mask(s) computed by Mask R-CNN~\protect\cite{he17}; \textbf{(c)} our resulting omnimatte (left: alpha matte, right: RGBA); \textbf{(d)} our estimated background layer. The bottom example (\textit{Bus Station}) shows a failure case: while the shadows are correctly associated with the people, the reflection cast on the window by the person in the top-right corner (marked by the red rectangle) is mistakenly grouped with the person in the bottom-left corner.
    }
    \label{fig:qualitative}\afterfigure
\end{figure*}

\subsection{Qualitative examples on real videos}
\label{sec:results_qual}
Figure~\ref{fig:qualitative} shows examples of our estimated omnimattes on a variety of real-world videos from DAVIS~\cite{davis2017}, CDW-2014~\cite{wang14cdw} (see Sec.~\ref{sec:bgsubtract}), and videos downloaded from YouTube. These examples span a wide range of dynamic subjects (e.g., people, animals or general moving objects such as a soccer ball), performing complex actions and generating various scene effects including shadows, reflections, water ripples,  dust and smoke.  None of the input object masks include these effects (see Fig.~\ref{fig:qualitative}(b)).

As seen in Fig.~\ref{fig:qualitative}(c-d) top, our method successfully associates the subjects with the scene effects that are related to them. In \emph{Blackswan}, the omnimatte of the swan captures its reflection and the water ripples it causes.  In  \emph{Elephant}, our omnimatte captures  the semi-transparent cloud of dust sprayed by the elephant, as well as the shadow the elephant casts on the ground.
In \emph{Tennis}, the running player casts thin shadows, which the omnimatte correctly separates from the shadows in the background. Additionally, although the player's racket is not included in the input mask (b), it is reconstructed in our omnimatte result; this  demonstrates our model's ability to reconstruct objects that are attached to the main subject even when given incomplete input masks.

In \emph{Soccer}, we show a two-subject example where our model estimates a separate omnimatte for each of the subjects: a person (top row) and a soccerball (bottom row). Our model successfully separates the person's shadow from that of the soccerball up until the final few frames of the video, where part of the person's shadow appears in the soccerball's omnimatte (full video in SM).

Another two-subject example is shown in \emph{Bus Station} where two people walk away from each other. Our model correctly associates
each person with their shadow in this challenging case.  However, the reflection cast on the window by the person on the right (top) is incorrectly placed in the left person's layer (bottom). 
The challenge of this scene lies in both the spatial proximity of the reflection to the incorrect person, 
and the similar motions (both people in the scene are moving consistently at the same speed).
Lu,~\el~\cite{lu2020layered} showed that layers tend to `grab' spatially proximal effects; in this case,
the reflection is actually closer to the person who is \textit{not} casting the reflection.
\cite{lu2020layered} additionally showed that correlated motions are grouped in the same layers; as both people
are walking in synchronization, the network places the reflection in the incorrect person's layer.

\begin{figure}[t!]
    \centering
    \vspace{-.1in}
    \includegraphics[width=\columnwidth]{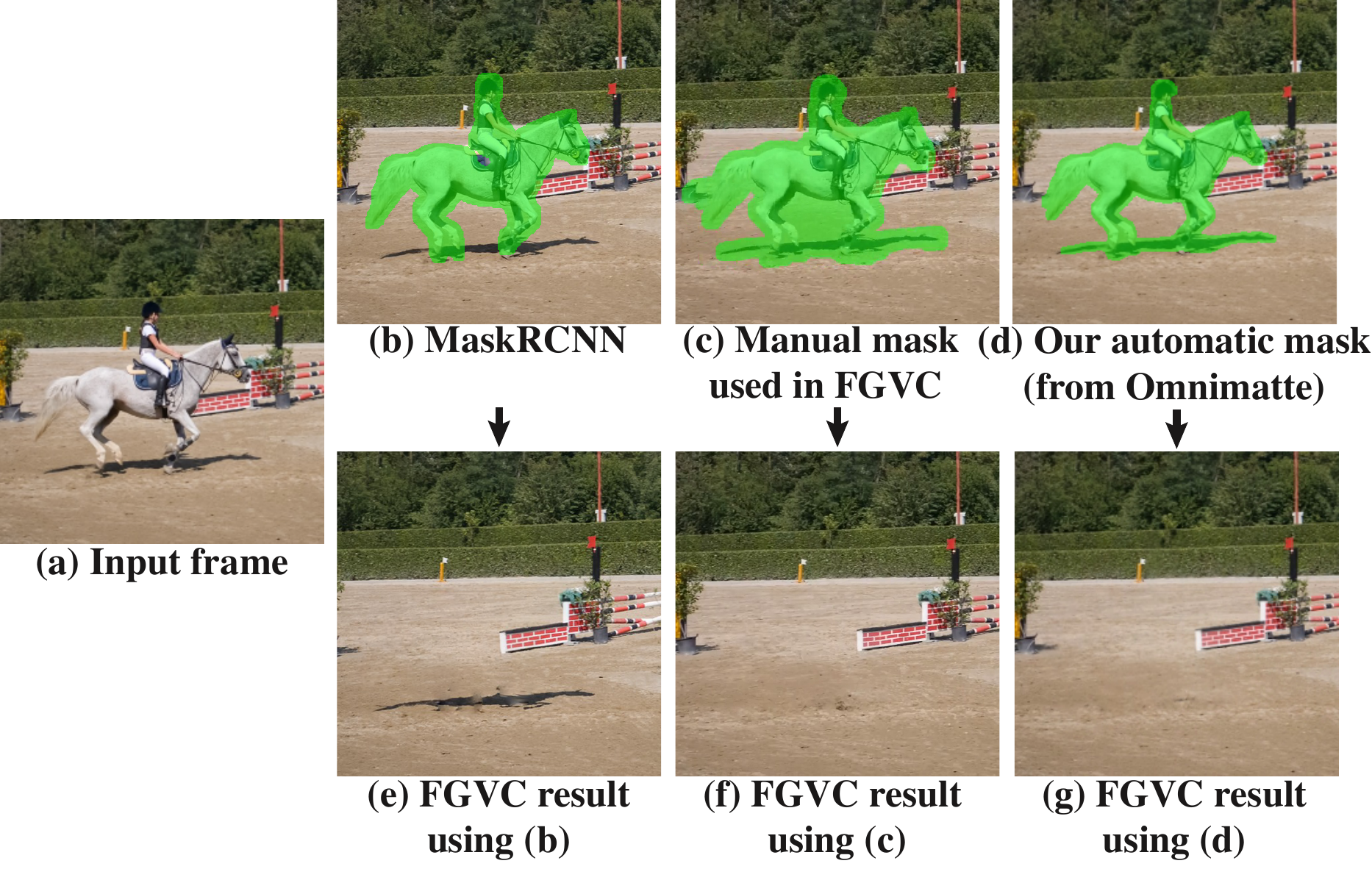}
    \caption{{\bf Omnimattes as input to state-of-the-art object removal.} Results by FGVC~\cite{Gao-ECCV-FGVC} using different types of input masks: \textbf{(b)} Raw masks from MaskRCNN do not capture shadows, and produce unrealistic results \textbf{(e)}. By using manually annotated masks that include the shadow \textbf{(b)}, both the horse and the shadow are removed \textbf{(f)}. \textbf{(d)} binary mask \emph{automatically derived from omnimatte} produces comparable result \textbf{(g)} when inputted to FGVC. 
    }
    \label{fig:vidcompmask}\afterfigure
\end{figure}

\vspace{-.1cm}
\subsection{Object Removal}
\vspace{-.1cm}
 Our method can be applied to remove a dynamic object from a video by either: (i) binarizing our omnimatte and using it as input to a separate video-completion method such as FGVC~\cite{Gao-ECCV-FGVC}, or by (ii)   simply excluding the object's omnimatte layer from our reconstruction.
 
 As shown in Fig.~\ref{fig:vidcompmask}(b,e), removing an object but not its correlated effects produces an unrealistic result (object removed but its shadow remains). Typically such effects are manually annotated to create a conservative binary mask of the regions to remove (Fig.~\ref{fig:vidcompmask}(c)). To show that an omnimatte can replace manual editing, we derive a binary mask by thresholding our soft alpha at $0.25$ and dilating by $20$ pixels, and inputting it to FGVC~\cite{Gao-ECCV-FGVC}. Fig.~\ref{fig:vidcompmask}(c) shows both the horse and its shadow are removed, demonstrating that our derived mask is comparable to a manually annotated mask.

\begin{figure}[t!]
\vspace{-.15in}
    \centering
    \includegraphics[width=\columnwidth]{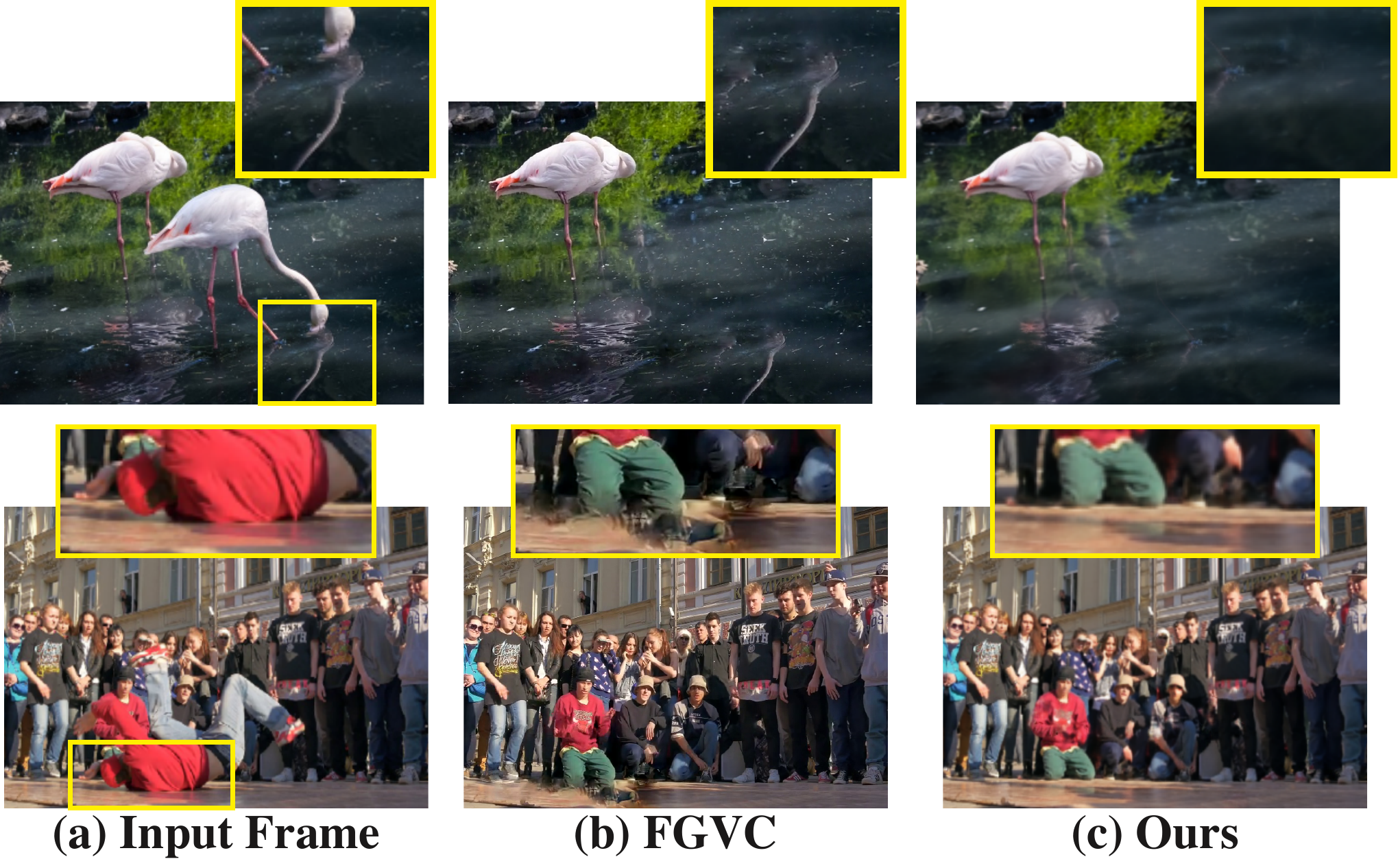}
    \caption{{\bf Direct omnimatte-based removal.} For each input frame \textbf{(a)} we show the result of removing a foreground object by excluding its omnimatte from the reconstruction \textbf{(c)}, compared to FGVC~\cite{Gao-ECCV-FGVC} \textbf{(b)}.}
    \label{fig:vidcomp}\afterfigure
\end{figure}

Fig.~\ref{fig:vidcomp} shows a comparison between omnimatte removal (approach (ii) above) and FGVC using manual masks.
In the \emph{flamingo} example, our method removes not only the flamingo but also its reflection in the water beneath it. FGVC relies on a mask that does not include the reflection, thus the reflection remains intact in their result. Our omnimattes bypass the need to manually label such semi-transparent effects.
In the \textit{breakdance} example, both the crowd and the dancer are moving. To handle this case, we assign an omnimatte to the dancer and a separate single omnimatte to the crowd. Fig.~\ref{fig:vidcomp}(c) shows the crowd omnimatte composited with the background layer.  The FGVC result (b) shows artifacts on the ground where the dancer is removed, whereas our result is seamless and realistic.

\vspace{-.1cm}
\subsection{Comparison with Shadow Detection}
\vspace{-.1cm}
We show qualitative comparisons with a recent state-of-the-art shadow detection method, ISD~\cite{wang2020isd}, a deep-learning based method that takes an RGB image as input and produces segments for object-shadow pairs. ISD integrates a MaskRCNN-like object detection stage (Detectron2~\cite{wu2019detectron2}), hence it does not require or allow an input mask. 

Figure~\ref{fig:shadow} compares our result with ISD on two challenging scenes, where a person casts a shadow onto another object (a bench), and where a person's shadow is occluded by another object (a dog). Our method successfully handles and outperforms ISD in both cases. Occlusions and shadows cast on other objects present particularly difficult cases for purely data-driven methods such as ISD, since the appearance of the shadow depends on the relative configuration of multiple objects in the scene, presenting a combinatorial explosion of scenarios for training. In contrast, our method analyzes and leverages space-time information throughout the entire video to perform these complex object-effects associations.

\begin{figure*}[]
\vspace{-.15in}
    \centering
    \includegraphics[width=\textwidth]{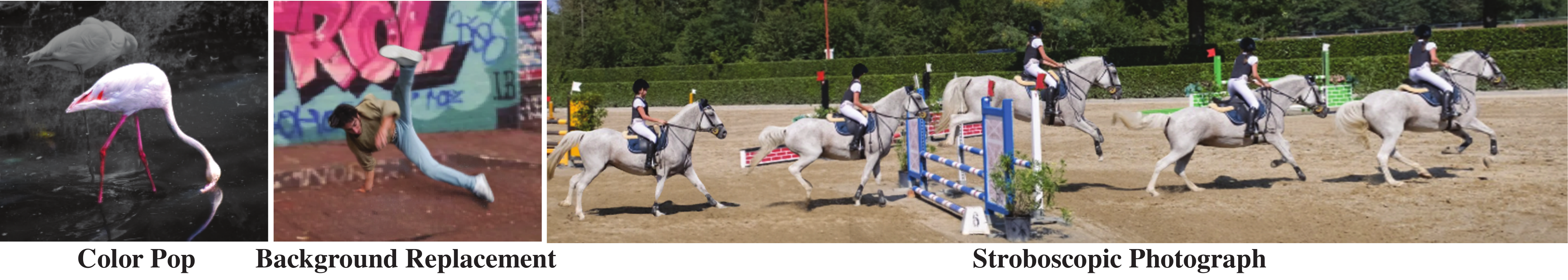}
    \vspace{-.2in}
    \caption{{\bf Video editing with Omnimattes.} Effects such as ``color pop'' (left; subject in color, background in grayscale), background replacement (center), and stroboscopic photography (right) all benefit from capturing the subjects' associated effects with an Omnimatte. Note the color pop present in the flamingo's reflection and the correct placement of shadows in the background replacement and stroboscopic photograph examples. See the SM for the full details and before/after videos.}
    \label{fig:effects}\afterfigure
\end{figure*}

\begin{figure}
    \centering
    \includegraphics[width=.9\columnwidth]{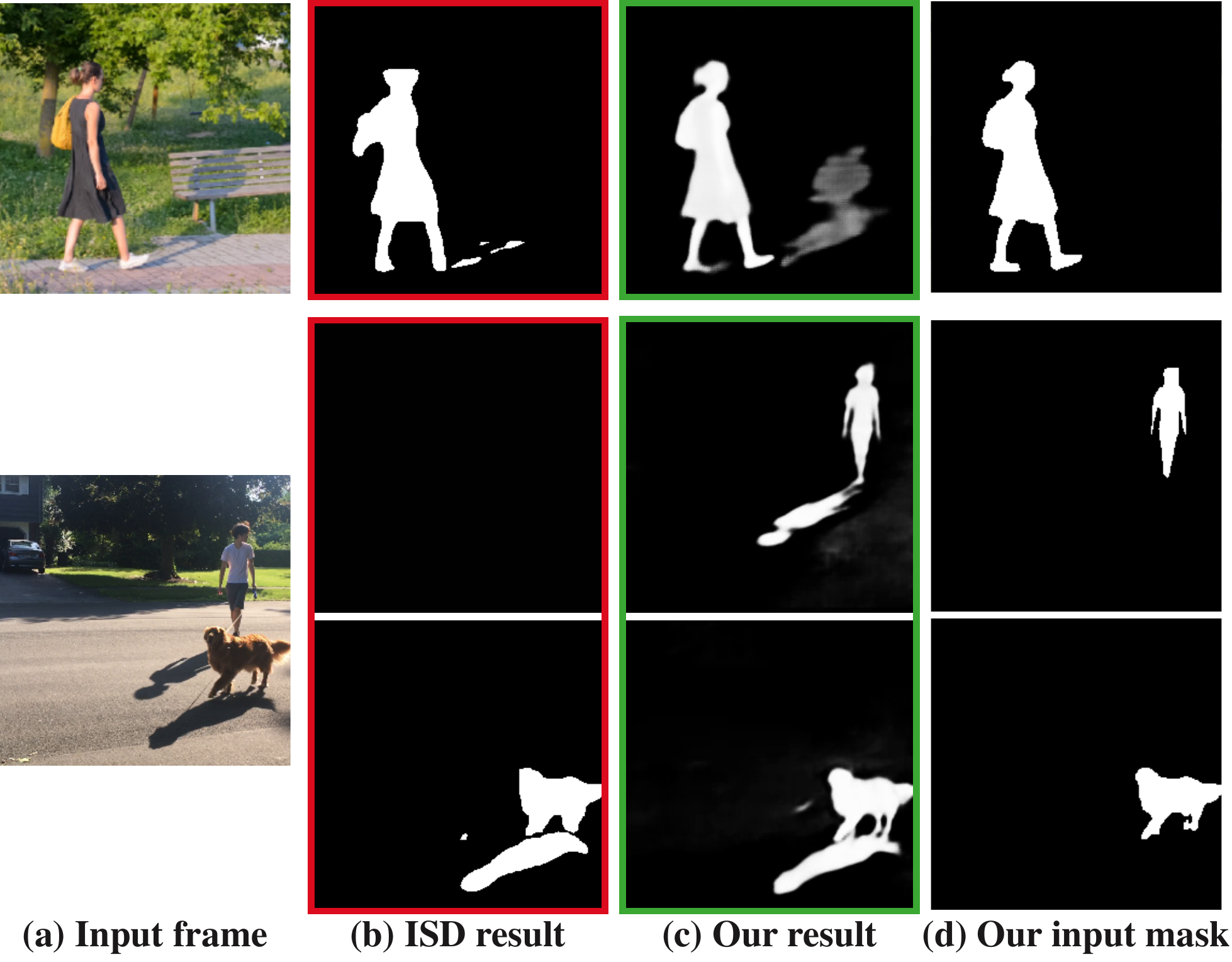}
    \caption{{\bf Comparison with shadow detection.} \textbf{(b)} Results produced by ISD~\protect\cite{wang2020isd}, a recent state of the art, single-image shadow detection method, and \textbf{(c)} our results when using \textbf{(d)} MaskRCNN masks as input.  See SM for additional comparisons.}

    \label{fig:shadow}\afterfigure
\end{figure}

\vspace{-.1cm}
\subsection{Comparison with Background Subtraction}
\begin{figure}[t]
    \centering
    \includegraphics[width=\columnwidth]{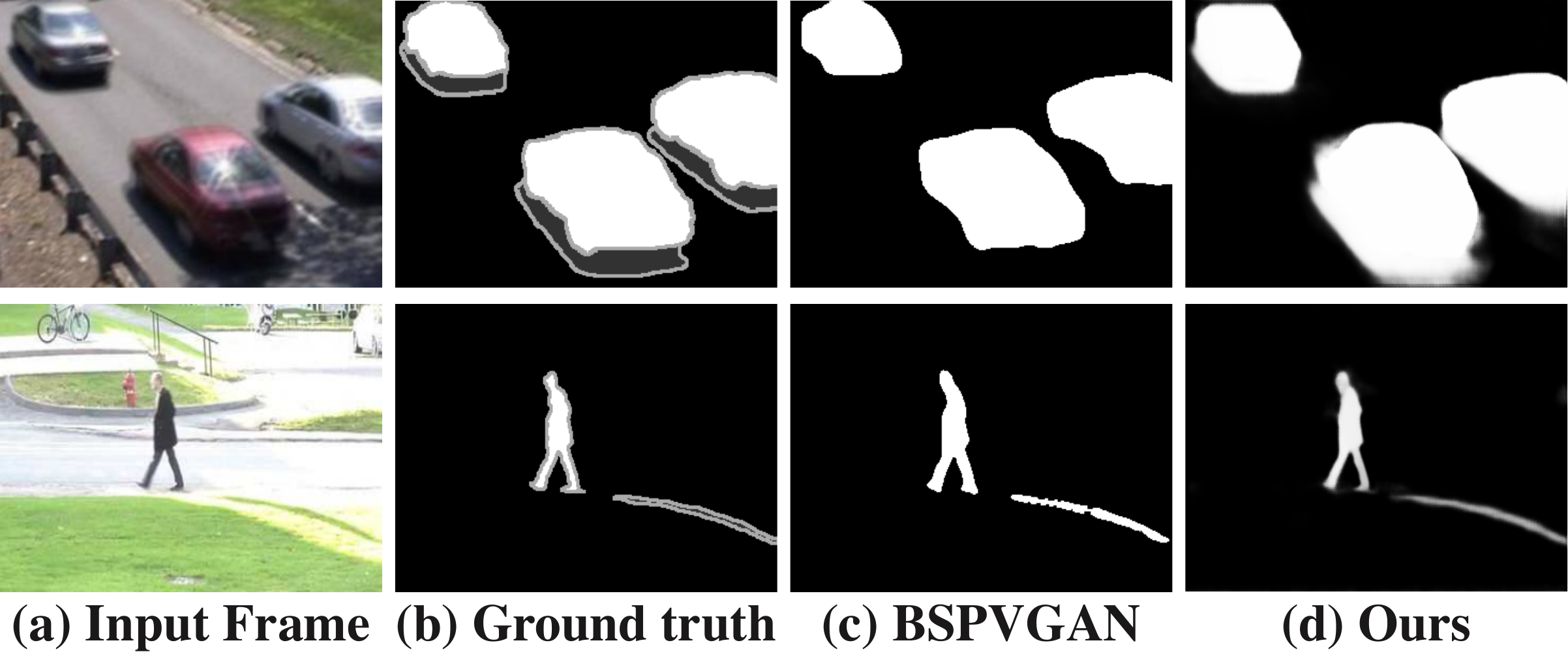}
    \caption{{\bf Comparison with background subtraction.} We used selected videos from the CDW-2014 change detection dataset~\cite{wang14cdw} (examples input frames in \textbf{(a)}), with ground truth, manually segmented objects and shadows (\textbf{b}, white pixels = moving objects, dark gray pixels = shadows, light gray pixels = `unknown', typically at boundaries). \textbf{(c)} Result by a top-performing (on this dataset) background subtraction method~\cite{zheng20}. \textbf{(d)} Our result (alpha mattes of estimated omnimattes). Numerical experiments are summarized in Table~\ref{tab:fgseg}. More results can be found in the SM.}
    \label{fig:cdw}
\end{figure}

\vspace{-.1cm}
\label{sec:bgsubtract}
We quantitatively evaluate our approach on the task of background subtraction using a change detection dataset, CDW-2014~\cite{wang14cdw}, which has ground-truth pixel-level labels for objects and hard shadows. We selected a subset of videos that contain objects and their shadows (see Fig.~\ref{fig:cdw} for sample frames and labels). 
We manually excluded ``bad weather'' and ``low framerate'' categories to avoid low quality videos, and selected short clips of up to 5 moving objects.
The selected subset contains 12 clips, each with 40 - 115 frames, for approximately 950 frames in total. 
While the background subtraction task requires only separating foreground from background, we demonstrate the additional capabilities of our method by also segmenting the effects for individual object instances.

We convert our soft omnimattes into a single, hard segmentation mask using a fixed threshold value and report the Jaccard index ($\mathcal{J}$) and Boundary measure ($\mathcal{F}$)~\cite{Perazzi2016} in Table~\ref{tab:fgseg}. We compare with two top-performing methods on CDW-2014, FgSegNet~\cite{lim2018learning} and BSPVGAN~\cite{zheng20}, which were trained on subsets of CDW-2014. Our method outperforms FgSegNet and matches the performance of BSPVGAN, despite not being trained supervised on CDW-2014.

\vspace{-.1cm}
\subsection{Comparison with Layered Neural Rendering}
\label{sec:retiming}
\vspace{-.15cm}
Fig.~\ref{fig:uvmethod} shows a qualitative comparison with the human-specific, layered neural rendering method by Lu~\el~\cite{lu2020layered}. In \cite{lu2020layered}, people are parameterized explicitly using per-frame UV maps that represent each individual's geometry, and a per-person trainable texture map that represents appearance. Instead, we use binary masks and pre-computed optical flow to represent object regions (see Sec.~\ref{sec:method}). For comparison, we used binary masks extracted from their UV maps.

In both examples, our method achieves comparable results to \cite{lu2020layered}, successfully capturing the trampoline deformations, shadows and reflections, yet with a generic, much simpler input. 
Note that the input masks derived from the UV maps provided by \cite{lu2020layered} represent the full body of a person even if they are occluded in the original frame. This allows our model to inpaint object regions and scene effects that are occluded in some frames but  visible in others, as in \cite{lu2020layered}. The ability of our model to inpaint occluded regions even when using incomplete masks is also evident in the person-dog example in Fig.~\ref{fig:teaser}, where the person and their shadow are reconstructed in our omnimatte.   However, we note that in cases where the input mask is substantially occluded, the output omnimatte will show occlusion as well; thus in order to deal with large occlusions, a full-object mask should be inputted to the model, as done in \cite{lu2020layered}.

\begin{table}
\small
\begin{tabular}{ c|c|c|c }
 Method & $\mathcal{J}$\&$\mathcal{F}$(Mean) $\uparrow$ & $\mathcal{J}$(Mean) $\uparrow$ & $\mathcal{F}$(Mean) $\uparrow$\\ 
 \hline
 FgSegNet~\cite{lim2018learning} & 0.675 & 0.631 & 0.719\\
 BSPVGAN~\cite{zheng20} & \textbf{0.756} & \textbf{0.718} & 0.793\\
 Ours & 0.754 & 0.711 & \textbf{0.797}
 \end{tabular}
\caption{We compare our method to the two top-performing methods on CDW-2014~\cite{wang14cdw}. We report the Jaccard index ($\mathcal{J}$) and Boundary measure ($\mathcal{F}$) on a subset of the data that includes objects and their shadows. Our method performs at par or better than the two background subtraction methods.}
\label{tab:fgseg} \afterfigure
\end{table}

\begin{figure*}[t!]
\label{fig:ablation}
\vspace{-.15in}
  \begin{minipage}[c]{0.25\textwidth}
        \caption{{\bf Ablations.} We ablate several components of our method: {\bf Left:} \textbf{(a-b)} sample input frame and our result using our full method, and \textbf{(c)} removing the brightness adjustment, and \textbf{(d)} removing the background offset (Section \ref{sec:camera}). \textbf{Right:} we show two examples comparing our method with and without the flow component \textbf{(f-g)}.}
    \end{minipage}
    \hfill
  \begin{minipage}[c]{0.72\textwidth}
    \includegraphics[width=\textwidth]{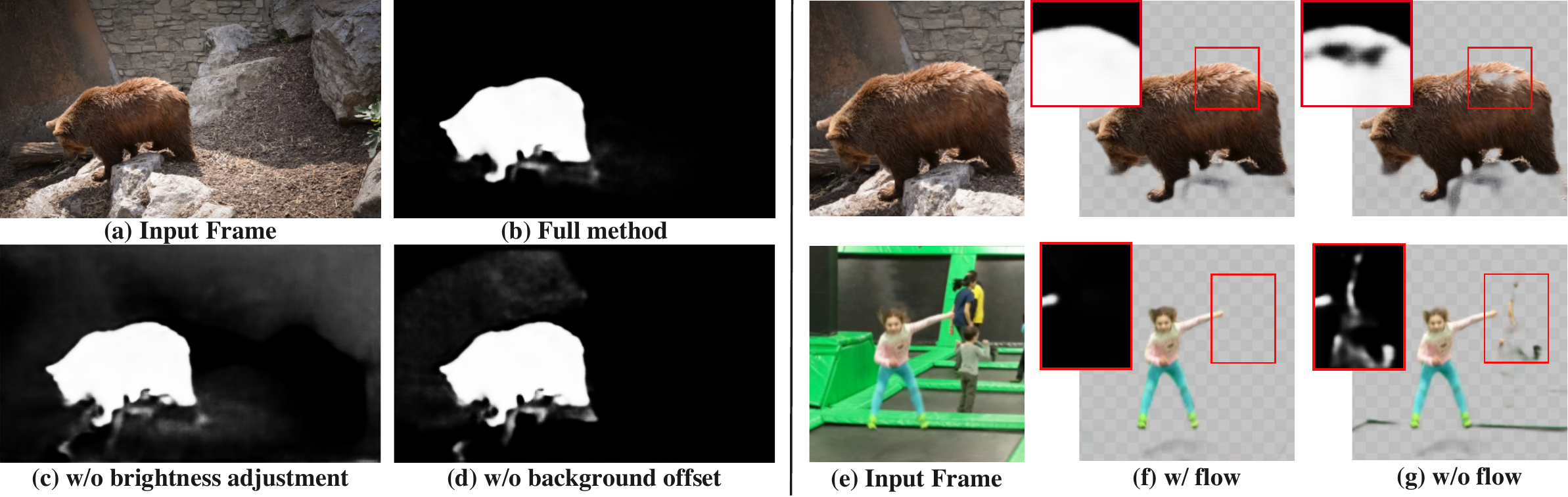}
  \end{minipage}
  \afterfigure
\end{figure*}

\begin{figure}
    \centering
    \includegraphics[width=\columnwidth]{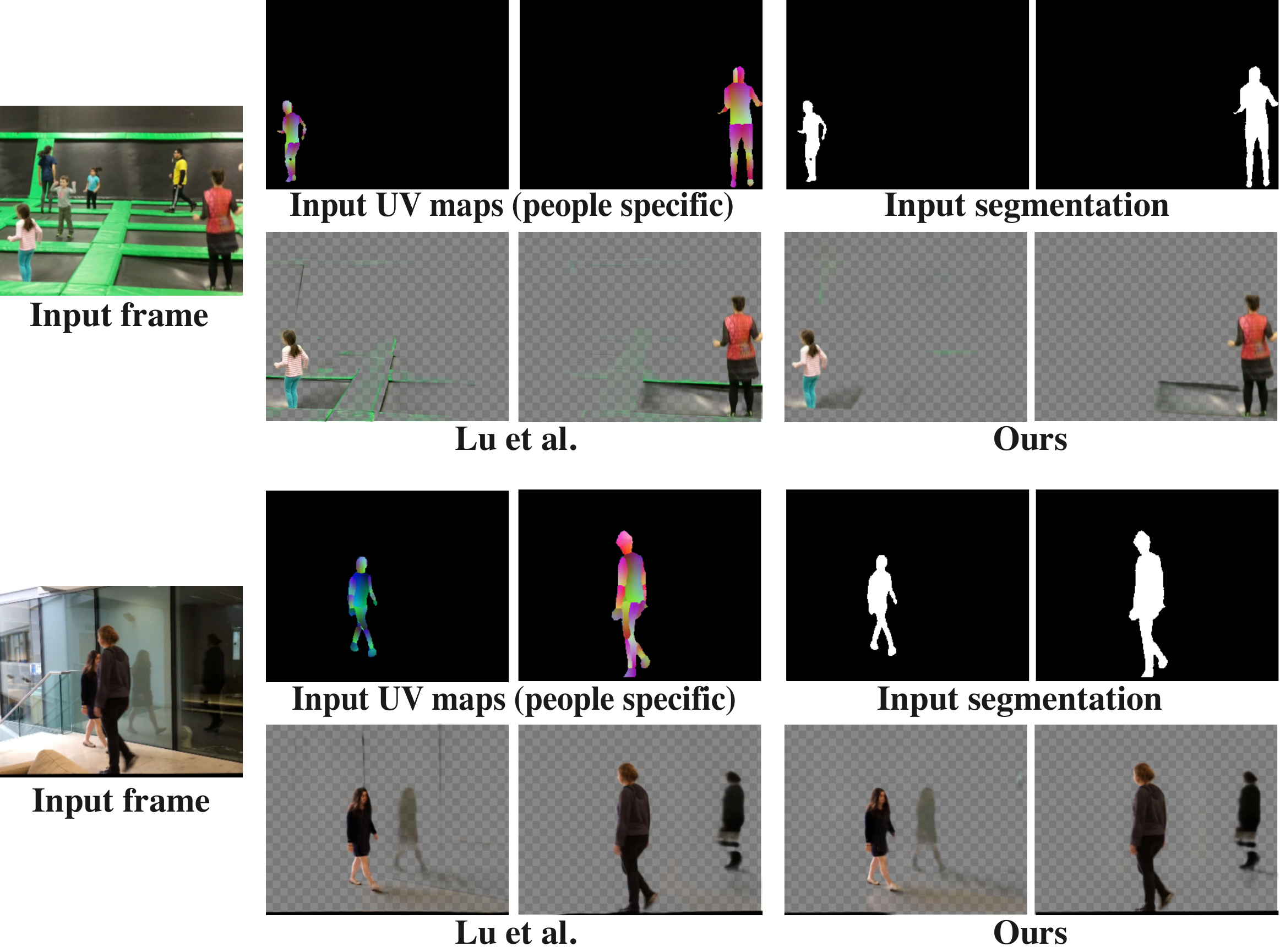}
    \caption{{\bf Comparison with Lu \etal~\cite{lu2020layered}.} We achieve comparable results to~\cite{lu2020layered} using just binary masks instead of the people-specific UV maps used in~\cite{lu2020layered}. These binary segmentation masks are easier to obtain and are \emph{general} -- allowing our method to support arbitrary objects (\cite{lu2020layered} is applicable just to people). Notice how our omnimattes capture trampoline deformation well (top two rows), and reflections in the glass (bottom two rows).}
    \label{fig:uvmethod} \afterfigure
\end{figure}

\vspace{-.1cm}
\subsection{Additional Video Editing Effects}
\vspace{-.15cm}
\label{sec:effects}
The additional information present in an omnimatte compared to a standard matte that includes only the subject allow simple creation of various video effects such as color pop, background replacement, or object duplication (Fig.~\ref{fig:effects}). Previously, creating these effects for videos containing shadows or reflections required extensive manual editing effort. Since the omnimatte is a standard RGBA image, these edits may be applied using standard video editing software. The omnimattes for color pop and background replacement were used unchanged, the horse jump alpha matte was adjusted with a linear contrast ramp. Please see SM for full details on creating these effects.

\vspace{-.1cm}
\subsection{Ablations}
\vspace{-.15cm}
In Fig. 9 \erika{fig 9 reference doesn't work} we ablate several components of our method.
Removing the brightness adjustment (c) and removing the background offset (d) both result in undesirable nonzero alpha values in the bear's omnimatte, due to lighting changes, vignetting, and homography inaccuracies that break the static background assumption  (Sec.~\ref{sec:camera}). Including both components results in a clean alpha matte (b).

We ablate the flow component of our model by removing both flow inputs and flow losses (Sec.~\ref{sec:flow_losses}), and show results in (g).
In the top row, part of the bear is missing from the omnimatte, and in the bottom row, the person's omnimatte incorrectly contains parts of other people. In contrast, our full model (f) has a complete bear and a clean person omnimatte.
These examples show that providing the model with motion information (flow) allows it to better associate scene elements with the correct objects, and 
prevents holes appearing in the foreground object’s alpha matte.

\vspace{-.1cm}
\subsection{Limitations}
\begin{figure}
    \centering
    \includegraphics[width=.9\columnwidth]{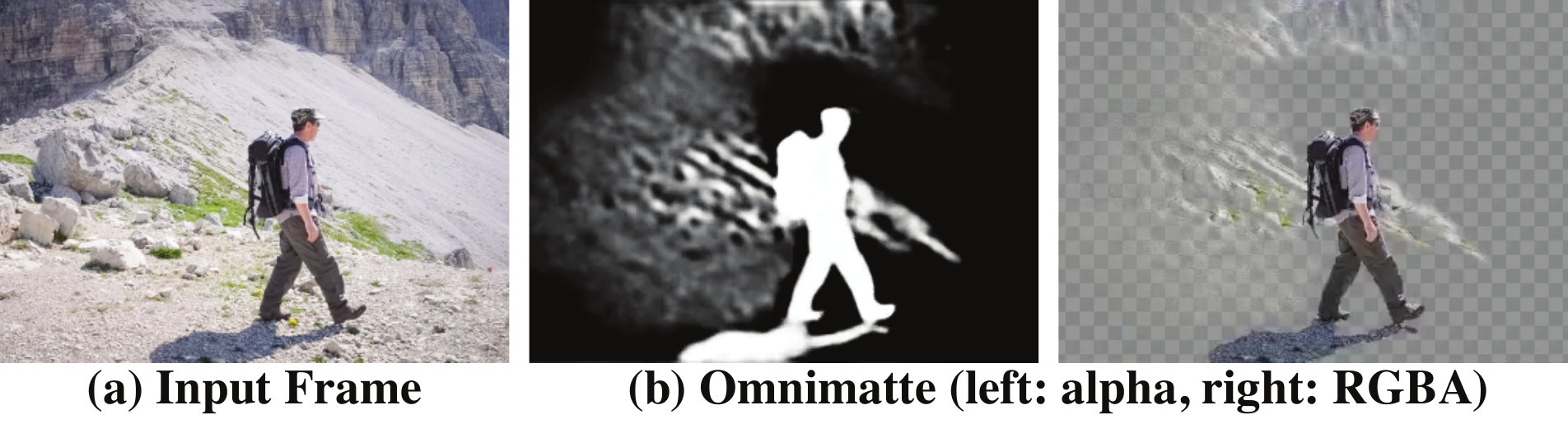}
    \caption{{\bf Failure case due to incorrect camera registration.} When the background motion cannot be accurately represented by a homography (in this case due to a significant depth variation in the scene), the predicted omnimatte may contain regions of the background to compensate for the registration inaccuracies.}
    \label{fig:failure} \afterfigure
\end{figure}

\vspace{-.15cm}
While our method allows for small deviations from a static background via smooth, coarse geometric and photometric offsets, when the homographies do not accurately represent the background, the omnimattes must correct for these errors by including background elements (e.g. rocks and grass, Fig.~\ref{fig:failure}). Conversely, we cannot separate objects or effects that remain entirely stationary relative to the background throughout the video. These issues could be addressed by building a background representation that explicitly models the 3D structure of the scene (e.g.~\cite{aliev2020npbg}). 

Finally, we observed that different random initializations of the network's weights may occasionally lead to different, sometimes undesirable, solutions (see supplemental material for visualization). %
We speculate that more reliable convergence could be obtained by further optimizing the order in which frames are introduced to the model.

\vspace{-.2cm}
\section{Conclusion}
\vspace{-.2cm}
We have posed a new problem: from an input video with one or more segmented moving subjects, we produce an \emph{omnimatte} for each subject -- an opacity map and color image that includes the subject itself along with the visual effects related to it. These effects can be reflections of the subject, shadows they cast, or attached objects. We have proposed a network and training framework for solving this new problem,
and have demonstrated omnimattes produced automatically for real-world videos with a variety of objects and associated effects. We have also shown how omnimattes can support a variety of video editing applications.

\vspace{-.5cm}
{\small
\paragraph{Acknowledgements.}
This work was supported in part by an Oxford-Google DeepMind Graduate Scholarship and a Royal Society Research Professorship. We thank Weidi Xie for assisting with object removal baselines.}

{\small
\bibliographystyle{ieee_fullname}
\bibliography{egbib}
}

\end{document}


\title{Omnimatte: Associating Objects and Their Effects in Video \\ {\large Supplementary Material} }  %

\maketitle
\thispagestyle{empty}
\pagenumbering{gobble}

\section{Formulation Details}

\subsection{Flow Reconstruction Weight (Section 3.3, Eq 6)}
The confidence at each pixel $p$ is defined as:
{\small \begin{equation}
    W_t(p) =  W_{t}^{lr}(p) \cdot W_{t}^{p}(p) \cdot M_t(p),
\end{equation}}
where $W_t^{lr}$ is computed using a standard left-right flow consistency error by: $W_t^{lr}(p) = \max (1-e_t^{lr}(p), 0)$, and $e_{lr}$ is the forward-backward flow error. $W_t^p$ measures photometric error and is given by: $W_t^p = \mathbbm{1}_{e_p<\beta}$, where $e_p = \|\textit{Warp}(I_{t+1}; F_{t,t+1}) - I_t\|_1$ measures the photometric difference between $I_{t}$ and $I_{t+1}$ when backward warped using the flow, and $\beta=20$. Finally, we mask the regions outside of the object mask $M_t$ since semi-transparent effects such as shadows tend to have inaccurate flow.

\subsection{Warp and Brightness Adjustment (Section 3.4)}
To compensate for minor camera stabilization errors, we apply the learnable warp at each pixel $\mathbf{x}$ of the background layer:
{\small \begin{equation}
    \hat{C}^0_t(\mathbf{x}) = C^0_t(\mathbf{G_w}[t,x_v,x_u])%
\end{equation}}
where $\mathbf{G_w}$ is a $n/10 \times 4 \times 7 \times 2$ grid of offset vectors, $n$ is the number of frames in the video, and $[\cdot,\cdot,\cdot]$ denotes trilinear filtering. 
We additionally apply a learnable coarse brightness scaling to the final composite:
{\small \begin{equation}
    \hat{Comp}(\mathcal{L}_t, o_t) = Comp(\mathcal{L}_t, o_t) \cdot \mathbf{G_b}[t,x_v,x_u]
\end{equation}}
where $\mathbf{G_b}$ is a $n/10\times 4 \times 7$ grid of brightness coefficients.
The values of $\mathbf{G_w}$ and $\mathbf{G_b}$ are optimized together with the network parameters. 

The effect of this adjustment layer can be seen in Fig.~\ref{fig:error_vs_alpha}. Adding the background adjustment significantly increases the sparsity of the alpha mattes for the same reconstruction quality. 

\subsection{RGBA Detail transfer (Section 3.5)}
We adopt the same detail transfer step as in Lu, \etal~\cite{lu2020layered}
to produce high-resolution omnimattes by transferring detail from the original frame to the CNN outputs.
We first compute the residual between the CNN output and the original frame, and
determine the amount of the residual to transfer to each RGBA layer using the transmittance map $\tau_t^i$ for layer $i$ at time $t$:
\begin{equation}
    \tau_t^i = 1.0 - \textit{Comp}_{\alpha}(\mathcal{L}_t \setminus \{ L_t^j \mid j < i \}, o_t \setminus \{ j \mid j < i\} )
\end{equation}
where $Comp_{\alpha}$ denotes the alpha channel of the composite produced by the network. The final layer colors with the transferred detail are:
\begin{equation}
    C_t^i = Cnr_t^i + \tau_t^i (I_t  - \textit{Comp}(\mathcal{L}_t, o_t))
\end{equation}
where $Cnr$ is the color produced by the network. 

\section{Implementation Details}

\paragraph{Network Architecture}
We adopt the same network architecture as in Lu, \etal~\cite{lu2020layered}, with the exception of replacing batchnorm with instance norm:
\begin{center}
\begin{tabular}{ |c|c|c|c|c| } 
 \hline
 & layer type(s) & channels & stride & activation\\
 \hline
 1 & conv & 64 & 2 & leaky\\
 2 & conv, IN & 128 & 2 & leaky\\
 3 & conv, IN & 256 & 2 & leaky\\
 4 & conv, IN & 256 & 2 & leaky\\
 5 & conv, IN & 256 & 2 & leaky\\
 6 & conv, IN & 256 & 1 & leaky\\
 7 & conv, IN & 256 & 1 & leaky\\
 8 & skip5, convt, IN & 256 & 2 & relu\\
 9 & skip4, convt, IN & 256 & 2 & relu\\
 10 & skip3, convt, IN & 128 & 2 & relu\\
 11 & skip2, convt, IN & 64 & 2 & relu\\
 12 & skip1, convt, IN & 64 & 2 & relu\\ 
 13 & conv & 4 & 1 & tanh\\
 \hline
\end{tabular}
\end{center}
`IN' refers to instance normalization, `convt' refers to convolutional transpose. All convolutions are $4 \times 4$.
Additionally, we have a convolutional layer following layer 12 which outputs 2 channels for optical flow.

\paragraph{Training Details} We implement our network in JAX~\cite{jax2018github} and Haiku~\cite{haiku2020github}. We optimize our full objective (Eq.~2,  Section 3.1), with relative weights: $\lambda_\text{r} = .005,     \lambda_\text{m} = 50$ until $\mathbf{E}_\text{mask}$ falls below 0.05, and is $0$ afterward, and $\lambda_\text{w} = .005$. 
We use the Adam optimizer~\cite{adamoptimizer} with an initial learning rate of 1e-3.
We train each video with a batch size of 32 on Google Cloud v3-8 TPU for 2000 epochs. We resize all videos to $256 \times 448$ spatial resolution. 
Training the `Bear' sequence from the DAVIS dataset~\cite{davis2017}, which has 82 frames and 1 output omnimatte, takes 2 hours.

\subsection{Video effects details (Section 4.6)}

All the video effects were created in \emph{postprocessing} in NUKE~\cite{nuke}, a standard video compositing package. That is, we produce the effects by editing our output omnimattes (and using the camera homographies), whereas the omnimatte network model itself is not used. 

\paragraph{Color Pop:} to create the ``color pop'' effect, we create two versions of the original video, one with color desaturated and one with color amplified. We blend the amplified version over the desaturated version using the alpha matte from the foreground layer.

\paragraph{Background Replacement:} replacing the background while preserving camera motion is accomplished by treating the new image as a new extended canvas. Thus, we create a new time-varying background frames $C_t^0$ by applying the original camera homographies.  %

\paragraph{Stroboscopic Photograph:} to create the background of the stroboscopic photograph, we apply the inverse camera homographies to the background color images $C_t^0$ and accumulate them into the canvas space with over compositing. This step creates a clean background canvas without the foreground objects. We then apply the same inverse transformations to the foreground layers and composite them on top. For the ``horsejump-low'' photo, we picked 15 frame intervals.

\begin{figure*}
    \centering
    \includegraphics[width=\textwidth]{figures/error_vs_alpha.png}
    \caption{{\bf Reconstruction error vs. matte sparsity for multiple runs on DAVIS.} Each graph measures the L1 reconstruction error against the fraction of pixels above 20\% alpha. Lower is better for both axes. A pixel above 20\% alpha is considered ``visible,'' so this value gives a measurement of amount of clutter in the matte. Each video is run with 5 random seeds for each of 3 ablation conditions: full method, no flow input or flow loss (``No Flow''), and no background offset or brightness adjustment (``No BG Adjust''). Note that ``No BG Adjust'' tends to produce more matte clutter, especially for videos with considerable camera motion (e.g. ``horsejump-low'', ``hike'').}
    \label{fig:error_vs_alpha}
\end{figure*}

\begin{figure*}[h]
    \centering
    \includegraphics[width=\linewidth]{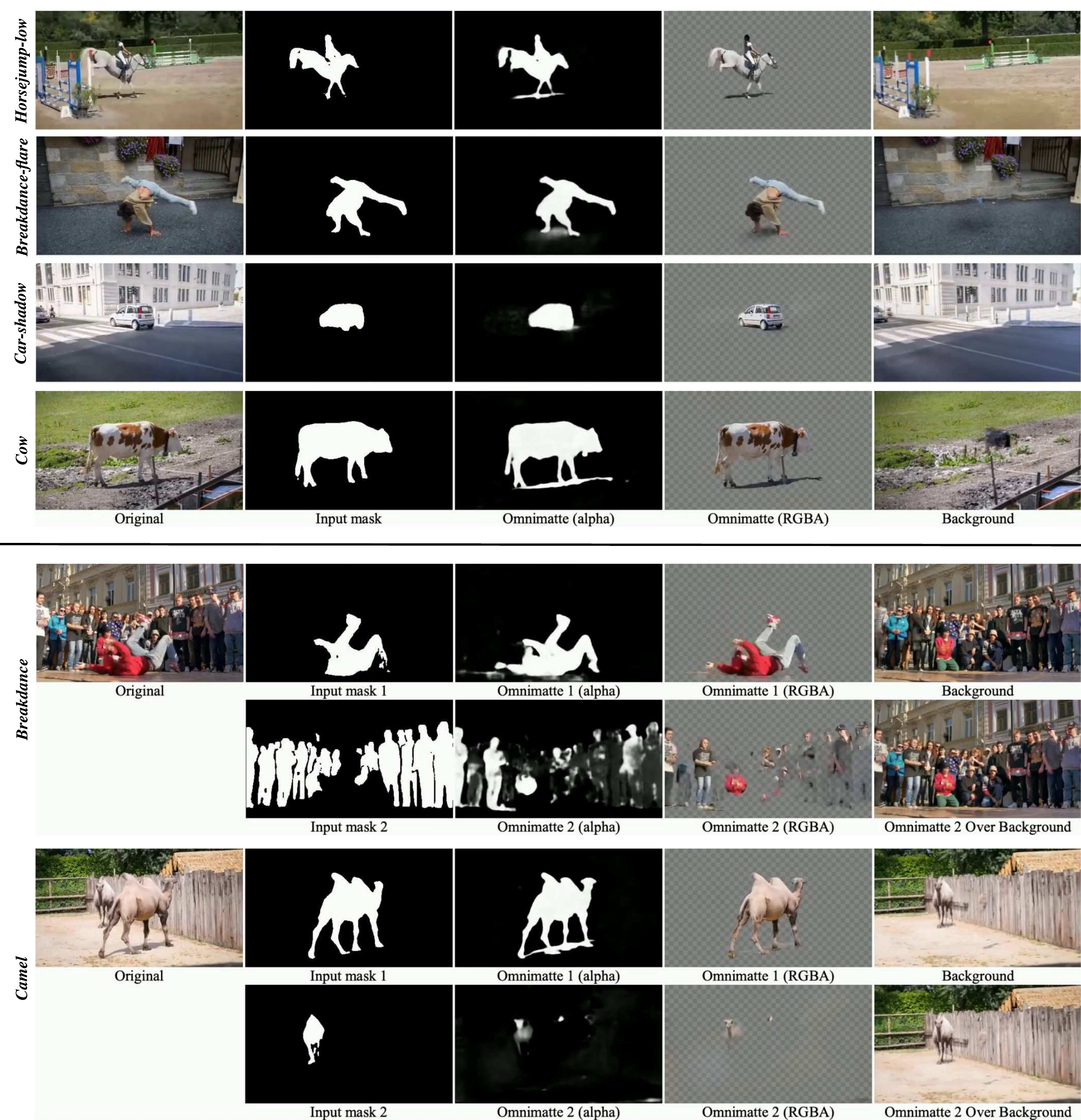}
    \caption{\textbf{More Omnimatte results on DAVIS~\cite{davis2017}.}
     We show, from left to right, the original RGB video frame, the input mask, the predicted omnimatte's alpha and RGBA visualizations, and the predicted background layer.
     Our method successfully detects shadows in each example, and can also
     handle inaccurate input masks (``Breakdance'', row 1: arm missing in input mask is included in prediction).
     In ``Camel'' row 2, the camel in the back is largely static, which results in it
     being primarily captured by the background layer.
    }
    \label{fig:supp1}
\end{figure*}

\begin{figure*}[h]
    \centering
    \includegraphics[width=\linewidth]{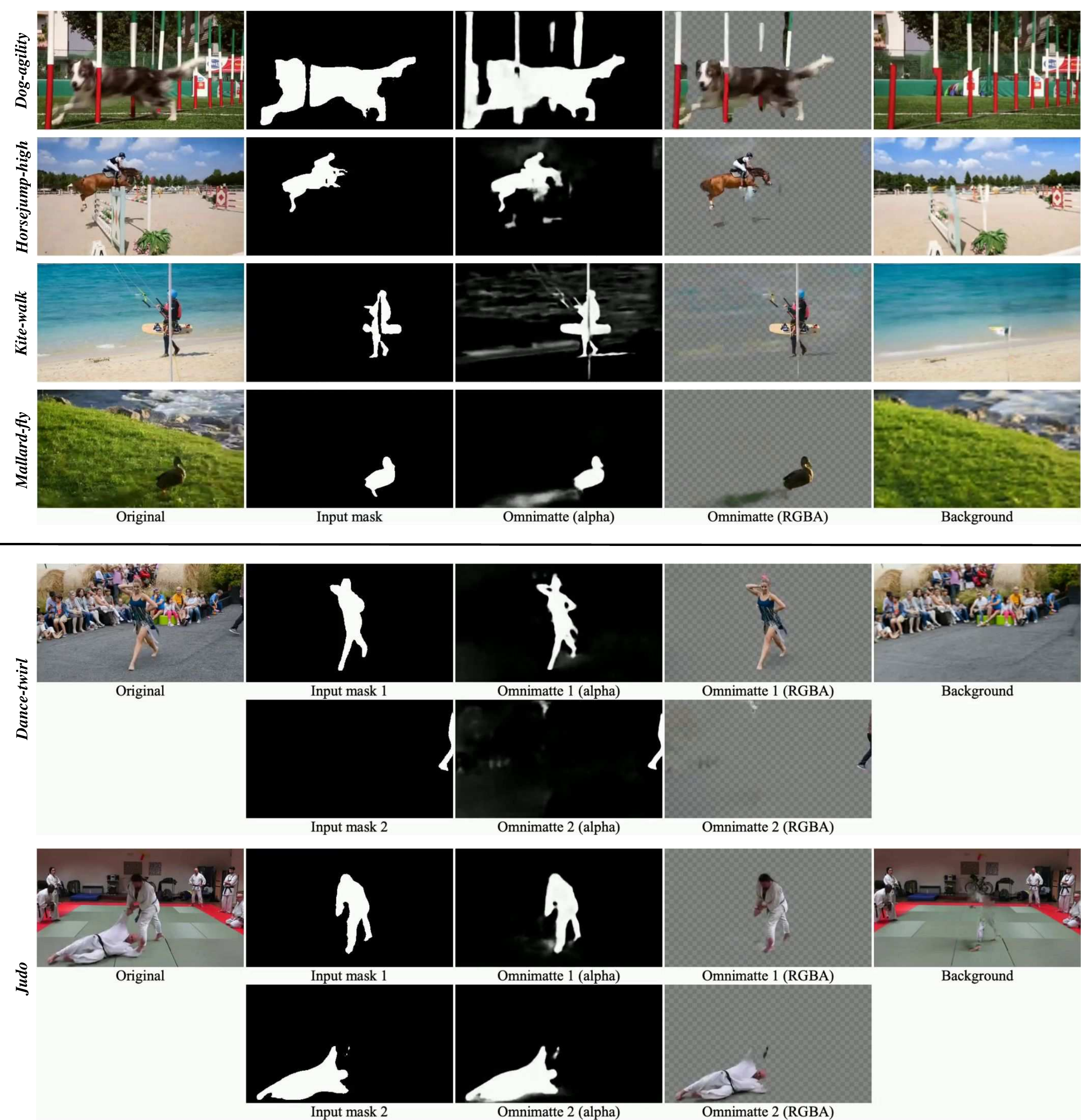}
    \caption{\textbf{More Omnimatte results on DAVIS~\cite{davis2017}.}
     We show, from left to right, the original RGB video frame, the input mask, the predicted omnimatte's alpha and RGBA visualizations, and the predicted background layer.
     We demonstrate our method on a variety of challenging scenes and effects, including shaking poles
     (``Dog-agility'') and detached shadows (``Horsejump-high'').
     Our method manages to capture loose clothing not present in the input mask (``Dance-twirl'').
     The ocean waves in ``Kite-walk'' are reconstructed in the foreground object layer because
     they cannot be represented by the static background layer.
    }
    \label{fig:supp2}
\end{figure*}

\begin{figure*}[h]
    \centering
    \includegraphics[width=\linewidth]{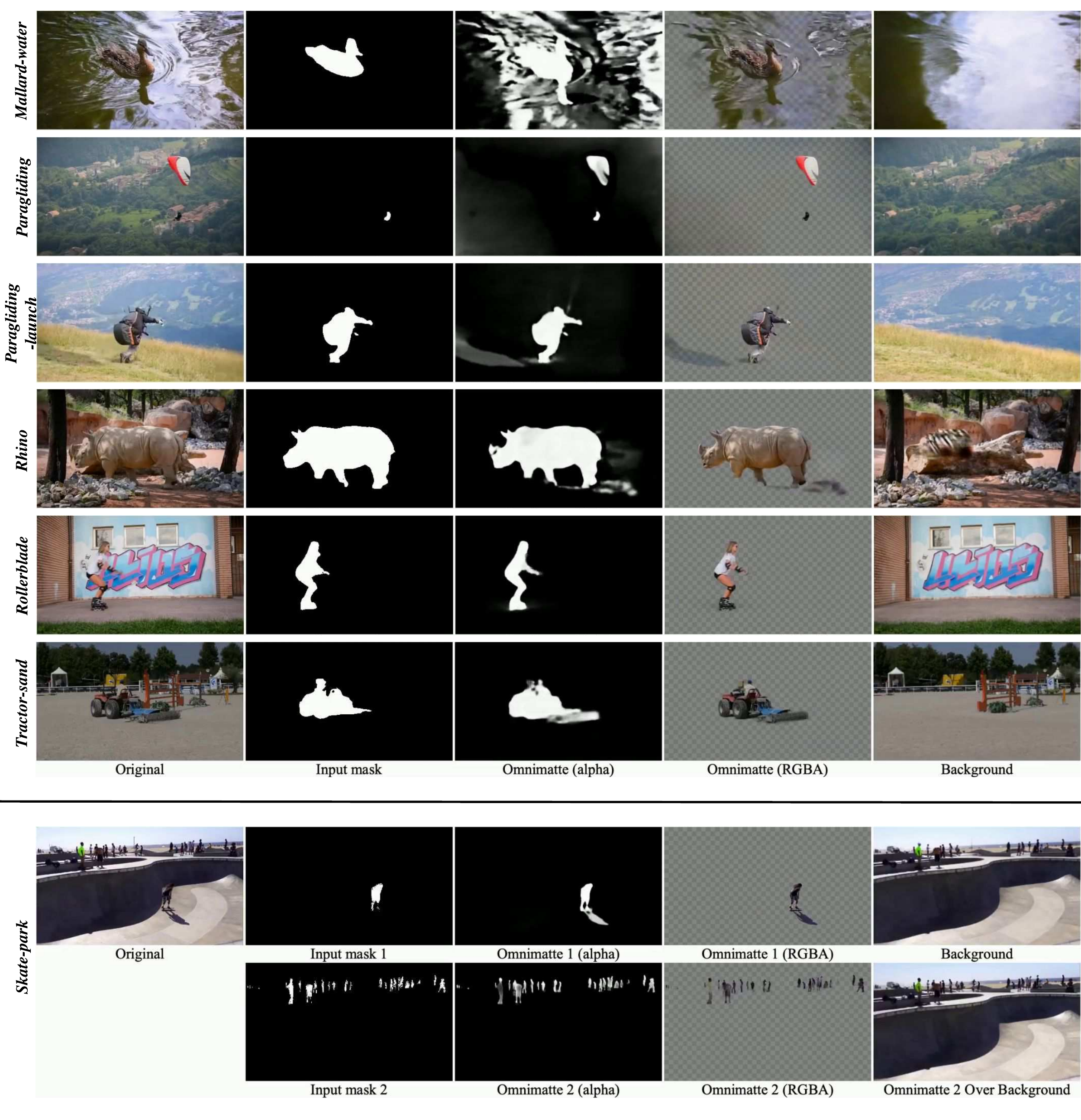}
    \caption{\textbf{More Omnimatte results on DAVIS~\cite{davis2017}.}
     We show, from left to right, the original RGB video frame, the input mask, the predicted omnimatte's alpha and RGBA visualizations, and the predicted background layer.
     Our method is able to capture large related objects or shadows (``Paragliding'', ``Paragliding-launch'').
     The water ripples in ``Mallard-water'' are reconstructed in the foreground object layer because
     they cannot be represented by the static background layer.
    }
    \label{fig:supp3}
\end{figure*}

\begin{figure*}[h]
    \centering
    \includegraphics[width=\linewidth]{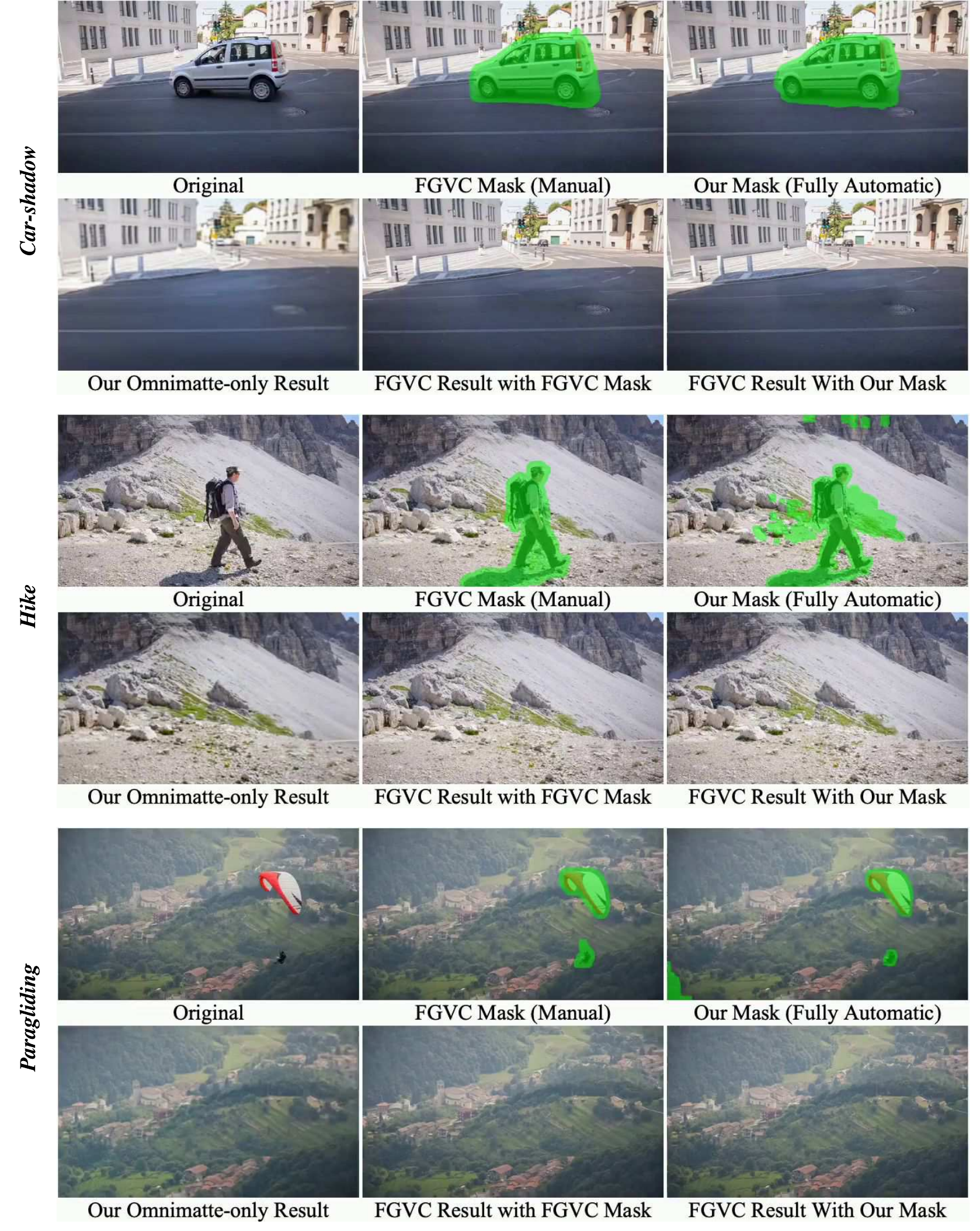}
    \caption{\textbf{More object removal comparisons with FGVC~\cite{Gao-ECCV-FGVC}.}
    For each example, we show in the top row from left to right: original RGB video frame,
    the input mask used for FGVC which was manually generated, and our input mask which was
    automatically generated by thresholding the predicted Omnimatte; in the bottom row:
    our object removal result by excluding the object layer from composition,
    the FGVC result using their manual input mask, and the FGVC result using our automatic input mask.
    We achieve comparable results using our automatic method.
    }
    \label{fig:supp4}
\end{figure*}

\begin{figure*}[h]
    \centering
    \includegraphics[width=\linewidth]{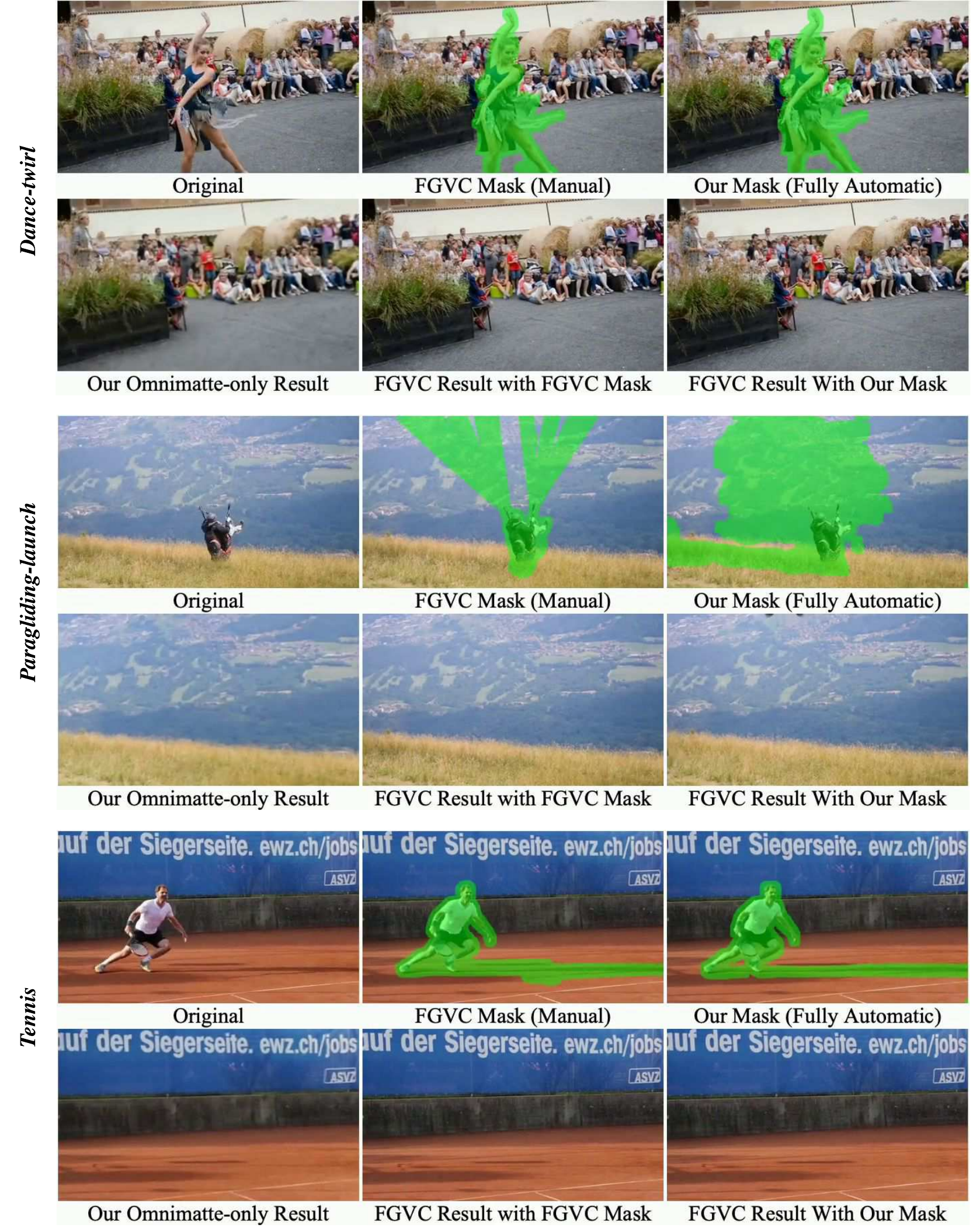}
    \caption{\textbf{More object removal comparisons with FGVC~\cite{Gao-ECCV-FGVC}.}
    For each example, we show in the top row from left to right: original RGB video frame,
    the input mask used for FGVC which was manually generated, and our input mask which was
    automatically generated by thresholding the predicted Omnimatte; in the bottom row:
    our object removal result by excluding the object layer from composition,
    the FGVC result using their manual input mask, and the FGVC result using our automatic input mask.
    We achieve comparable results using our automatic method.
    }
    \label{fig:supp5}
\end{figure*}

\begin{figure*}[h]
    \centering
    \includegraphics[width=\linewidth]{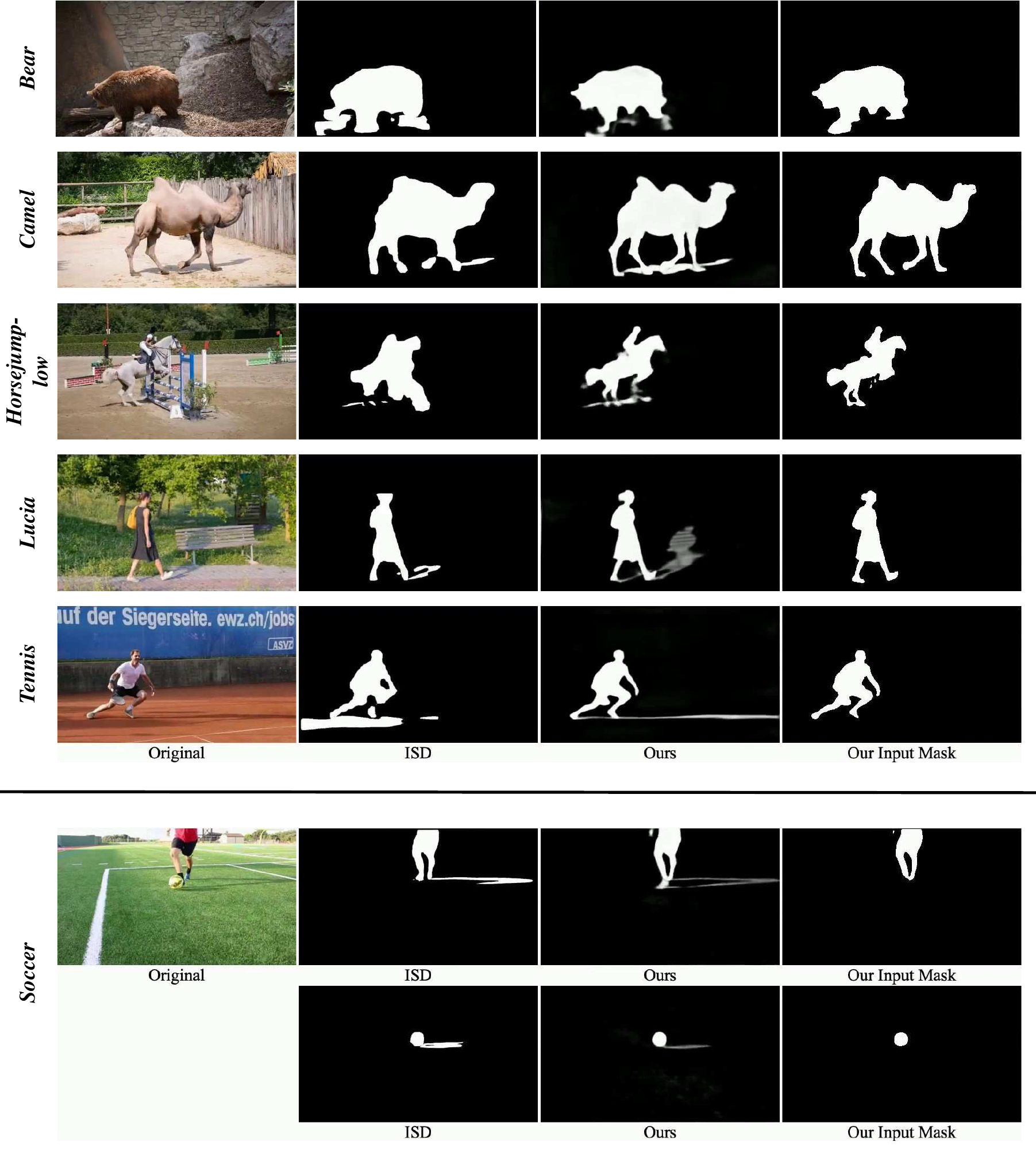}
    \caption{\textbf{More shadow detection comparisons with ISD~\cite{wang2020isd}.}
    For each example, we show from left to right: the original RGB video frame,
    the object+shadow prediction from ISD, our predicted Omnimatte alpha channel,
    and our input mask.
    Our method is able to take advantage of motion in the video, which often enables it
    to detect shadows more accurately than the single-image, data-driven method, ISD
    (e.g. in ``Tennis'', ISD mistakes other shadows on the ground for the person's shadow,
    but our method detects the correct shadow after observing the motion in the scene).
    }
    \label{fig:supp6}
\end{figure*}

\begin{figure*}[h]
    \centering
    \includegraphics[width=\linewidth]{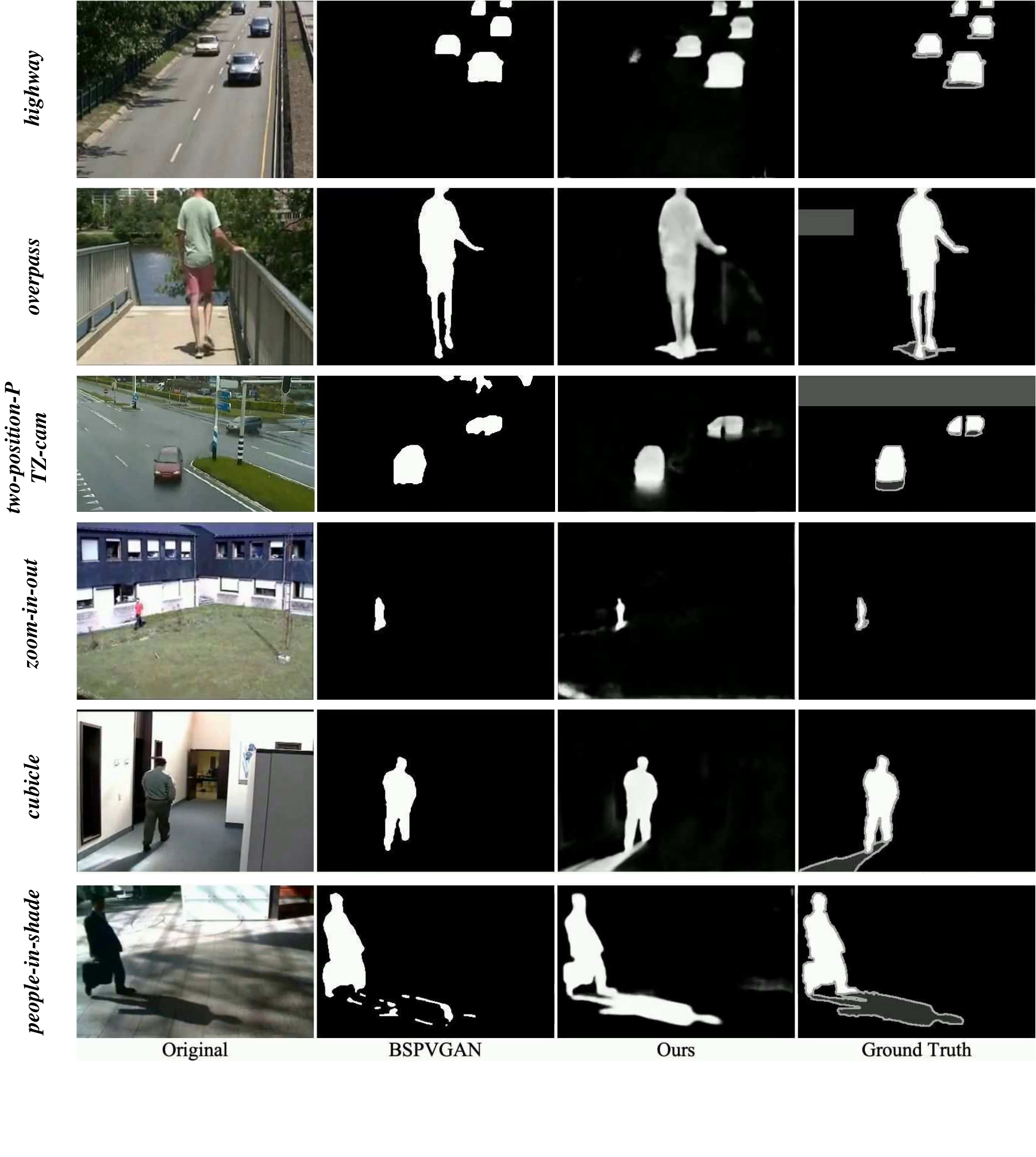}
    \caption{\textbf{More change detection comparisons with BSPVGAN~\cite{zheng20}.}
    We show comparisons on our subset of CDW-2014~\cite{wang14cdw}.
    For each example, we show from left to right: the original RGB video frame,
    the prediction from BSPVGAN, our predicted Omnimatte alpha channel,
    and the ground truth mask.
    The ground truth label definitions are as follows: white pixels = moving objects,
    dark gray pixels = shadows, light gray pixels = ‘unknown’ (typically at boundaries or unannotated regions).
    Our method is more successful at detecting shadows.
    }
    \label{fig:supp7}
\end{figure*}

{\small
\bibliographystyle{ieee_fullname}
\bibliography{egbib}
}